\documentclass{article}

\usepackage[preprint]{neurips_2021}

\usepackage[utf8]{inputenc} % allow utf-8 input
\usepackage[T1]{fontenc}    % use 8-bit T1 fonts
\usepackage{hyperref}       % hyperlinks
\hypersetup{colorlinks=true, linkcolor=blue, filecolor=magenta, urlcolor=magenta}
\usepackage{url}            % simple URL typesetting
\usepackage{booktabs}       % professional-quality tables
\usepackage{amsfonts}       % blackboard math symbols
\usepackage{nicefrac}       % compact symbols for 1/2, etc.
\usepackage{microtype}      % microtypography
\usepackage{xcolor}         % colors
\usepackage{bm}

\usepackage{multirow}  		% \multirow
\usepackage{amsmath} 		% \operatorname
\usepackage{wrapfig} 		% wraptable wrapfigure
\usepackage{graphicx} 		% figure
\usepackage{subcaption} 	% sub table
\title{3D Siamese Voxel-to-BEV Tracker for Sparse Point Clouds}
\author{%Le Hui, Jia Yuan, Mingmei Cheng, Jin Xie\thanks{Corresponding authors}, Xiaoya Zhang, Jian Yang$^{*}$
  Le Hui$^{\dagger}$, Lingpeng Wang$^{\dagger}$, Mingmei Cheng, Jin Xie$^{*}$, Jian Yang$^{*}$ \\
  PCA Lab, Nanjing University of Science and Technology, China\\
  \texttt{\{le.hui, cslpwang, chengmm, csjxie, csjyang\}@njust.edu.cn} \\
}
\begin{document}

\maketitle
\let\thefootnote\relax\footnotetext{$^{\dagger}$Equal Contributions, $^*$Corresponding authors.}
\let\thefootnote\relax\footnotetext{Le Hui, Lingpeng Wang, Mingmei Cheng, Jin Xie, and Jian Yang are with PCA Lab, Key Lab of Intelligent Perception and Systems for High-Dimensional Information of Ministry of Education, and Jiangsu Key Lab of Image and Video Understanding for Social Security, School of Computer Science and Engineering, Nanjing University of Science and Technology, China.}

\begin{abstract}
3D object tracking in point clouds is still a challenging problem due to the sparsity of LiDAR points in dynamic environments. In this work, we propose a Siamese voxel-to-BEV tracker, which can significantly improve the tracking performance in sparse 3D point clouds. Specifically, it consists of a Siamese shape-aware feature learning network and a voxel-to-BEV target localization network. The Siamese shape-aware feature learning network can capture 3D shape information of the object to learn the discriminative features of the object so that the potential target from the background in sparse point clouds can be identified. To this end, we first perform template feature embedding to embed the template's feature into the potential target and then generate a dense 3D shape to characterize the shape information of the potential target. For localizing the tracked target, the voxel-to-BEV target localization network regresses the target's 2D center and the $z$-axis center from the dense bird's eye view (BEV) feature map in an anchor-free manner. Concretely, we compress the voxelized point cloud along $z$-axis through max pooling to obtain a dense BEV feature map, where the regression of the 2D center and the $z$-axis center can be performed more effectively. Extensive evaluation on the KITTI and nuScenes datasets shows that our method significantly outperforms the current state-of-the-art methods by a large margin. Code is available at \url{https://github.com/fpthink/V2B}.

\end{abstract}

\section{Introduction}
Object tracking is an essential task in computer vision and has been widely in various applications, such as autonomous vehicle, mobile robotics, and augmented reality. In the past few years, many efforts~\cite{kristan2015visual,bertinetto2016fully,danelljan2017eco,valmadre2017end} have been made on 2D object tracking from RGB data. Recently, with the development of 3D sensor such as LiDAR and Kinect, 3D object tracking~\cite{Lebeda20142DON,Whelan2016ElasticFusionRD,Rnz2017CofusionRS,Kart2018HowTM,Liu2019ContextAwareTM} has attracted more attention. Lately, some pioneering works~\cite{gordon2004beyond,Feng2020ANO,Qi2020P2BPN} have focused on point cloud based 3D object tracking. However, due to the sparsity of 3D point clouds, 3D object tracking on point clouds is still a challenging task.

Few works are dedicated to 3D single object tracking (SOT) with only point clouds. As a pioneer, SC3D~\cite{giancola2019leveraging} is the first 3D Siamese tracker that performs matching between the template and candidate 3D target proposals generated by Kalman filtering~\cite{gordon2004beyond}. %Feng~\emph{et al.}~\cite{Feng2020ANO} followed SC3D and proposed a two-stage framework called Re-Track, which re-track the lost objects of the coarse stage in the fine stage. Moreover, they also use a shape completion network to regularize feature learning on the candidate target proposals. Since Kalman filtering is used to generate candidate 3D target proposals, they cannot perform end-to-end training, and consume much time when matching plenty of candidate target proposals.
%Furthermore, a shape completion network is used to regularize feature learning on the candidate target proposals. However, SC3D cannot perform end-to-end training and consumes much time when matching plenty of candidate target proposals, so it cannot be applied to real-time tracking.
Furthermore, a shape completion network is used to enhance shape information of candidate proposals in sparse point clouds, thereby improving the accuracy of matching. However, SC3D cannot perform the end-to-end training, and consumes much time when matching exhaustive candidate proposals.
%Since Kalman filtering is used to generate candidate target proposals, SC3D cannot perform end-to-end training, and consume much time when matching plenty of proposals.
Towards these concerns, Qi~\emph{et al.}~\cite{Qi2020P2BPN} proposed an end-to-end framework termed P2B, which first localizes potential target centers in the search area via Hough voting~\cite{qi2019deep}, and then aggregates vote clusters to generate target proposals. %Lately, Feng~\emph{et al.}~\cite{Feng2020ANO} proposed a two-stage framework called Re-Track, which re-track the lost objects of the coarse stage in the fine stage. Like SC3D, Re-Track adopts a shape completion network to regularize feature learning on the candidate target proposals.
Nonetheless, when facing sparse scenes, P2B may not be able to track the object accurately, or even lose the tracked object. On the one hand, it adopts random sampling to generate initial seed points, which further exacerbates the sparsity of point clouds. On the other hand, it is difficult to generate high-quality target proposals on sparse 3D point clouds. Although SC3D has enhanced shape information of candidate proposals, the low-quality candidate proposals obtained from sparse point clouds still degrade tacking performance.
%only uses a simple PointNet~\cite{qi2017pointnet} as the encoder that individually treats each point. Therefore, it ignores the local geometric structure, which makes it unable to learn discriminative features in sparse point clouds.
%To tackle above issues, we propose a novel Siamese point cloud tracker, which aims to improve the tracking performance in sparse scenes. Specifically, our intuition consists of three levels. Firstly, by strengthening the correlation learning between the template and the search area, we can provide rich target clues for the potential object in the sparse search area. Secondly, enhancing shape information in the sparse point cloud will provide discriminative information for distinguish the target from the background. Thirdly, we require a denser feature map to enhance the detection accuracy of the target center.

\begin{wrapfigure}{r}{0.45\textwidth}
	\vspace{-8pt}
	\centering
	\includegraphics[width=0.45\textwidth]{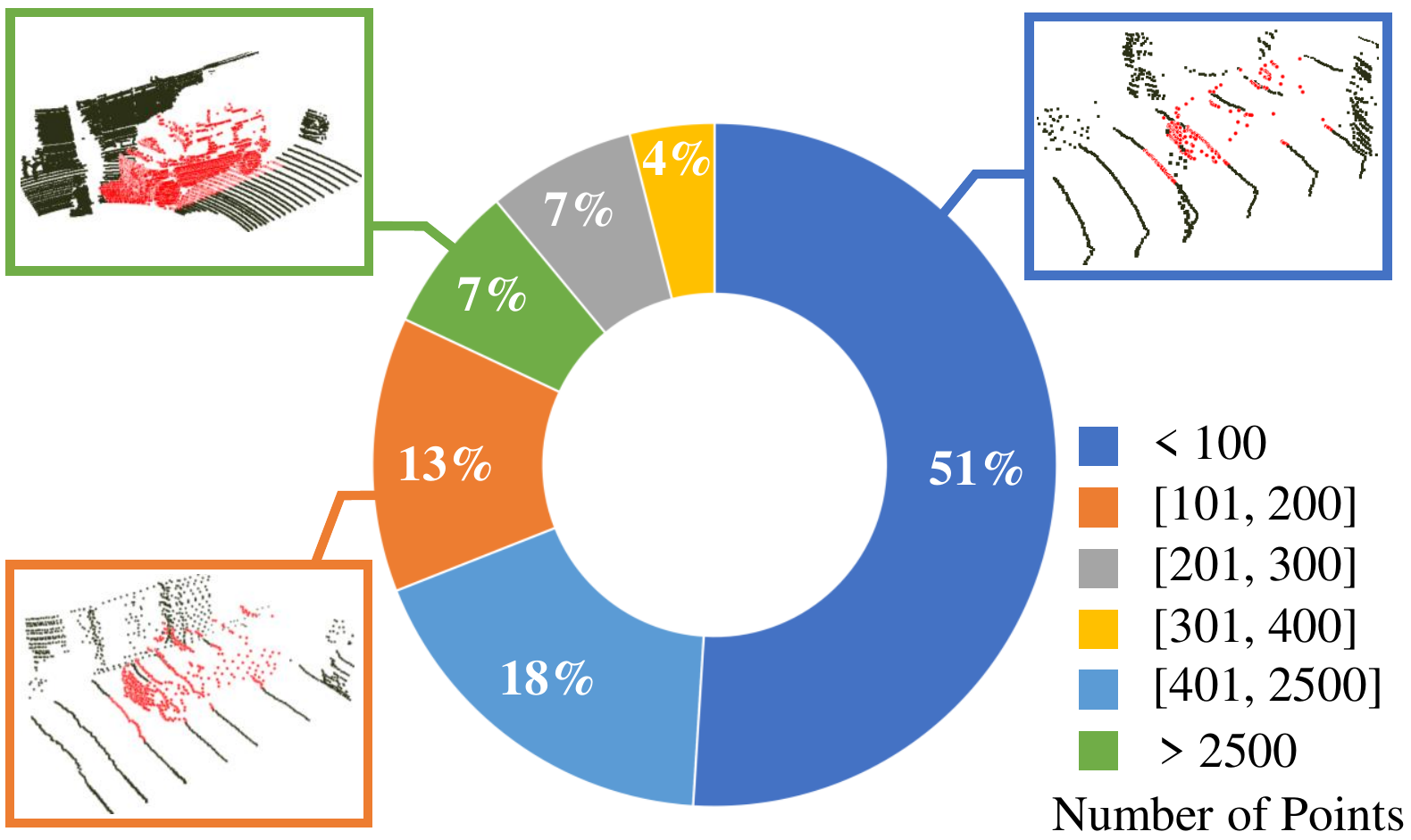}
	\vspace{-10pt}
	\caption{Statistics of the number of points on KITTI's cars. Cars are colored in \color{red}red.}
	\vspace{-8pt}
	\label{fig:motivation}
\end{wrapfigure}
As shown in Fig.~\ref{fig:motivation}, we count the number of points on KITTI's cars. It can be found that 51\% of cars have less than 100 points, and only 7\% of cars have more than 2500 points. When facing sparse point clouds, it is difficult to distinguish the target from the background due to the sparsity of point clouds. Therefore, how to improve the tracking performance in sparse scenes should be considered. Our intuition consists of two folds. First, enhancing shape information of target will provide discriminative information to distinguish the target from the background, especially in sparse point clouds. Second, due to the sparsity of the point cloud, it is difficult to regress the target center in 3D space. We hence consider compressing the sparse 3D space into a dense 2D space, and perform center regression in the dense 2D space to improve tracking performance.

In this paper, we propose a novel Siamese voxel-to-BEV (V2B) tracker, which aims to improve the tracking performance of 3D single object tracking, especially in sparse point clouds. We illustrate our framework in Fig.~\ref{fig:framework}. We first feed the template and search area into the Siamese network to extract point features, respectively. Then, we employ the global and local template feature embedding to strengthen the correlation between the template and search area so that the potential target in the search area can be effectively localized. After that, we introduce a shape-aware feature learning module to learn the dense geometric features of the potential target, where the complete and dense point clouds of the target are generated. Thus, the geometric structures of the potential target can be captured better so that the potential target can be effectively distinguished from the background in the search area. Finally, we develop a voxel-to-BEV target localization network to localize the target in the search area. In order to avoid using the low-quality proposals on sparse point clouds for target center prediction, we directly regress the 3D center of the target with the highest response in the dense bird's eye view (BEV) feature map, where the dense BEV feature map is generated by voxelizing the learned dense geometric features and performing max-pooling along the $z$ axis. Thus, with the constructed dense BEV feature map, for sparse point clouds, our method can more accurately localize the target center without any proposal.

In summary, we propose a novel Siamese voxel-to-BEV tracker, which can significantly improve tracking performance, especially in sparse point clouds. We develop a Siamese shape-aware feature learning network that can introduce shape information to enhance the discrimination of the potential target in the search area. We develop a voxel-to-BEV target localization network, which can accurately detect the 3D target's center in the dense BEV space compared to sparse 3D space. Extensive results show that our method has achieved new state-of-the-art results on the KITTI dataset~\cite{Geiger2012AreWR}, and has a good generalization ability on the nuScenes~\cite{nuscenes2019} dataset.

\section{Related Work}
\textbf{2D object tracking.} Numerous schemes~\cite{bromley1993signature,gordon2004beyond,kristan2016novel,wu2013online,bertinetto2016staple} have been presented and achieved impressive results in 2D object tracking. Early works are mainly based on correlation filtering. As a pioneer, MOSSE~\cite{bolme2010visual} presents stable correlation filters for visual tracking. After that, correlation-based methods use Circulant matrices~\cite{henriques2012exploiting}, kernelized correlation filters~\cite{henriques2014high}, continuous convolution filters~\cite{danelljan2016beyond}, factorized convolution operators~\cite{danelljan2017eco} to improve tracking performance. In recent years, Siamese-based methods~\cite{fan2019siamese,held2016learning} have been more popular in the tracking field. In~\cite{bertinetto2016fully}, Bertinetto~\emph{et al.} proposed SiamFC, a pioneering work that combines naive feature correlation with a fully-convolutional Siamese network for object tracking. %Subsequently, SiamRPN~\cite{li2018high} and SiamRPN++~\cite{li2019siamrpn++} improves the tracking performance by integrating the Siamese networks with RPN~\cite{ren2015faster} and deeper architectures, respectively.
Subsequently, some improvements~\cite{zhu2018distractor,wang2019fast,xu2020siamfc++,zhang2020ocean,yan2020alpha} are made to Siamese trackers, such as combining with a region proposal network~\cite{Feichtenhofer2017DetectTT,li2018high,Yu2020DeformableSA,Voigtlaender2020SiamRV} or an anchor-free FCOS detector~\cite{Choi2020VisualTB}, using a deeper architecture~\cite{li2019siamrpn++} or two-branch structure~\cite{he2018twofold}, exploiting attention~\cite{wang2018learning,Zhou2019SiamManSM} or self-attention~\cite{chen2021transformer}, applying triplet loss~\cite{dong2018triplet}. However, these methods are specially designed for 2D object tracking, so they cannot be directly applied to 3D point clouds. %Inspired by~\cite{bertinetto2016fully,Feichtenhofer2017DetectTT}, we develop a Siamese tracking-by-detection framework for 3D object tracking task. %in the case of sparse point clouds.

\textbf{3D single object tracking.} %Different from 2D object tracking, 3D object tracking leverages the geometric information in the 3D space to track objects instead of relying on the 2D appearance properties of the objects.
Early 3D single object tracking (SOT) methods focus on RGB-D information. As a pioneer, Song~\emph{et al.}~\cite{song2013tracking} first proposed a unified 100 RGB-D video dataset, which opened up a new research direction for RGB-D tracking~\cite{bibi20163d,liu2018context,Kart2019ObjectTB}. Based on RGB-D information, 3D SOT methods~\cite{spinello2010layered,luber2011people,Rnz2017CofusionRS} usually combines techniques from the 2D tracking with additional depth information. However, RGB-D tracking also relies on RGB information, and it may fail when the RGB information is degraded. Recent efforts~\cite{luo2018fast,wang2020pointtracknet} begin to use LiDAR point clouds for 3D single object tracking. Among them, SC3D~\cite{giancola2019leveraging} is the first 3D Siamese tracker, but it is not and end-to-end framework. Following it, Re-Track~\cite{Feng2020ANO} is a two-stage framework that re-tracks the lost objects of the coarse stage in the fine stage. Lately, Qi~\emph{et al.}~\cite{Qi2020P2BPN} proposed P2B, which solves the problem that SC3D cannot perform end-to-end training and consumes a lot of time. P2B adopts the tracking-by-detection scheme, which uses VoteNet~\cite{qi2019deep} to generate proposals and selects the proposal with the highest score as the target. Based on P2B, to handle sparse and incomplete target shapes, BAT~\cite{zheng2021box} introduces a box-aware feature module to enhance the correlation learning between template and search area. Nonetheless, when facing very sparse scenarios, VoteNet used in P2B and BAT may be difficult to generate high-quality proposals, resulting in performance degradation.

%Nonetheless, P2B ignores the effect of point cloud sparsity on tracking, which motivates us research.
%However, it is hard to accurately detect the 3D center under very sparse point clouds. To this end, we consider decomposing 3D target center into 2D center plus additional height so that each item can be regressed effectively and accurately.

%Recent efforts~\cite{luo2018fast,wang2020pointtracknet,giancola2019leveraging,Qi2020P2BPN,Feng2020ANO} begin to use LiDAR point clouds for 3D object tracking. Among them, SC3D~\cite{giancola2019leveraging} is the first 3D Siamese tracker for point clouds. SC3D first uses Kalman filtering~\cite{gordon2004beyond} to generate a lot of candidate 3D target proposals, and then performs tracking by matching the template area with the candidate proposals. To enhance semantic discrimination and geometric information, a shape completion network is used to regularize the feature learning. Feng~\emph{et al.}~\cite{Feng2020ANO} followed SC3D and proposed a two-stage framework called Re-Track, which re-track the lost objects of the coarse stage in the fine stage. SC3D and Re-Track cannot perform end-to-end training, and 3D search using Kalman filtering will consume much time. To this end, Qi~\emph{et al.}~\cite{Qi2020P2BPN} proposed a point-to-box (P2B) network that first uses Hough voting~\cite{qi2019deep} to cluster potential target centers and then generate target proposals without the time-consuming space search in~\cite{giancola2019leveraging}. However, 

\textbf{3D multi-object tracking.} Most 3D multi-object tracking (MOT) systems follow the same schemes with the 2D multi-object tracking systems, but the only difference is that 2D detection methods are replaced by 3D detection methods. Most 3D MOT methods~\cite{wu20213d,shenoi2020jrmot,kim2021eagermot} usually adopt tracking-by-detection schemes. Specifically, they first use a 3D object detector~\cite{shi2019pointrcnn,shi2020points,shi2020pv} to detect numerous objects of each frame, and then exploit the data association between detection results of two frames to match the corresponding objects. To exploit the data association, early works~\cite{scheidegger2018mono} use handcrafted features such as spatial distance. Instead, modern 3D trackers use motion information that can be obtained by 3D Kalman filters~\cite{patil2019h3d,chiu2020probabilistic,weng20203d} and learned deep features~\cite{yin2021center,zhang2019robust}.
%The data association usually indicates motion information
%Based on 3D Kalman filters~\cite{chiu2020probabilistic,weng20203d}, 3D trackers still can achieve impressive results by using the three-dimensional motion in a scene.
%Dedicated 3D trackers based on 3D Kalman filters~\cite{chiu2020probabilistic,weng20203d}
%most 3D multi-object tracking methods~\cite{wu20213d,patil2019h3d,chiu2020probabilistic,weng20203d,yin2021center,zhang2019robust,shenoi2020jrmot,kim2021eagermot} usually adopt tracking-by-detection schemes. Specifically, they first use a 3D object detector~\cite{shi2019pointrcnn,shi2020points,shi2020pv} to detect numerous objects of each frame, and then exploit the data association between detection results of two frames to match the corresponding objects. In addition, 3D multi-object tracking needs to consider track initiation, re-identification, and termination during tracking.

\textbf{Deep learning on point clouds.} With the introduction of PointNet~\cite{qi2017pointnet}, 3D deep learning on point clouds has stimulated the interest of researchers. Existing methods can be mainly divided into: point-based~\cite{qi2017pointnet++,pointcnn2018,kpconv2019,cheng2020cascaded}, volumetric-based~\cite{qi2016volumetric,liu2019point}, graph-based~\cite{wang2018dynamic,landrieu2018large,landrieu2019point,gacnet2019,cheng2021sspc,hui2021superpoint}, and view-based~\cite{su2015multi,su20153d,yang2019learning} methods. However, volumetric-based and view-based methods lose fine-grained geometric information due to voxelization and projection, while graph-based methods are not suitable for sparse point clouds since few points cannot provide sufficient local geometric information for constructing a graph. Thus, existing 3D tracking networks~\cite{giancola2019leveraging,Feng2020ANO,Qi2020P2BPN,zheng2021box} are point-based methods.

\section{Method}
Our work is specifically designed for 3D single object tracking in sparse point clouds. An overview of our framework is depicted in Fig.~\ref{fig:framework}. We first present the Siamese shape-aware feature learning network to enhance the discrimination of the potential target in search area (Sec.~\ref{sec:shape_learning}). We then localize the target by voxel-to-BEV target localization network (Sec.~\ref{sec:center_detection}).

\vspace{-5pt}
\subsection{Siamese Shape-Aware Feature Learning Network}\label{sec:shape_learning}
%In this section, we aim to learn dense geometric features of the potential target with sparse points in the search area. We first employ template feature embedding to encode the search area by learning the similarity of the global shape and local geometric structures between the template and search area. We then employ shape-aware feature learning to learn dense geometric features of the target by generating a dense and complete point cloud of the target.

%Due to the sparse and incomplete point clouds of the potential target in the search area, we then employ shape-aware feature learning to learn dense geometric features of the target, where the dense and complete point clouds of the target can be obtained. It is expected that the learned features from the generated dense point clouds can characterize the geometric structures of the target better.

%For 3D object tracking, it is crucial to establish a target-to-target relationship between the template and search area in order to better localize the potential target in the search area.
%The two branches both use PointNet++~\cite{qi2017pointnet++} as backbone and share parameters.
%Here, we perform template feature embedding to establish a correlation between the template and search area in order to better localize the potential target in the search area. 

\subsubsection{Template Feature Embedding}
Suppose the size of the template is $N$, and the size of the search area is $M$ (generally, $M > N$). Before template feature embedding, we first use the Siamese network to extract point features of the template and search area, denoted by $P=\{\bm{p}_i\}_{i=1}^{N}$ and $Q=\{\bm{q}_j\}_{j=1}^{M}$. The Siamese network consists of template branch and detection branch. In order to reduce the inference time of the network, we just use PointNet++~\cite{qi2017pointnet++} as the backbone and share parameters. It can be replaced with a powerful network such as KPConv~\cite{kpconv2019}. We then employ template feature embedding to encode the search area by learning the similarity of the global shape and local geometric structures between the template and search area. The illustration of template feature embedding is shown in the left half of Fig.~\ref{fig:siamese_shape}.

\begin{figure}
	%\vspace{-15pt}
	\centering
	\includegraphics[width=1.0\textwidth]{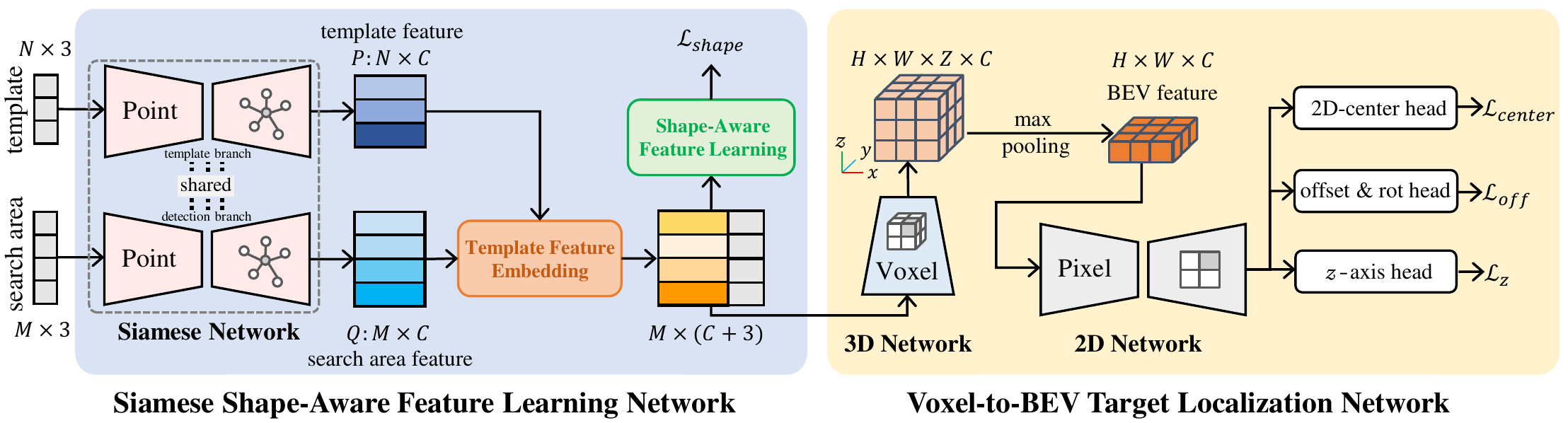}
	%\vspace{-15pt}
	\caption{The architecture of V2B. Given a template and search area, we first use Siamese network to obtain template and search area features. We then perform template feature embedding and shape-aware feature learning to enhance the ability to distinguish the target from the background. Finally, we perform voxel-to-BEV target localization to detect the 3D object center from the BEV.}
	\vspace{-15pt}
	\label{fig:framework}
\end{figure}

\textbf{Template global feature embedding.} We use the multi-layer perceptron (MLP) network to adaptively learn the correlation between the template and search area. The similarity between the template and search area is formulated as:
\begin{equation}
\setlength{\abovedisplayskip}{1pt}
\setlength{\belowdisplayskip}{1pt}
\bm{w}_{ij} = f_{corr}(\bm{p}_i, \bm{q}_j) = \operatorname{MLP}(\bm{p}_i-\bm{q}_j), \forall \bm{p}_i\in P, \bm{q}_j\in Q
\end{equation}
where $\bm{p}_i-\bm{q}_j$ characterizes the difference between the two feature vectors and $\bm{w}_{ij}\in\mathbb{R}^C$ is the correlation weight between two points. The global shape information of the template is given by:
%into the search area, and the embedded function $f_{emb}$ is written as:
\begin{equation}
\setlength{\abovedisplayskip}{1pt}
\setlength{\belowdisplayskip}{1pt}
\bm{q}^{'}_{j} = f_{emb}(\bm{q}_j, \bm{p}_1,\bm{p}_2, \ldots, \bm{p}_N)=\operatorname{MLP}(\underset{i=1,2,\ldots,N}{\operatorname{MAX}}\{\bm{p}_i\cdot\bm{w}_{ij}\}), \forall \bm{q}_j\in Q
%\vspace{-5pt}
\end{equation}
where $\operatorname{MAX}$ represents the max pooling function and $\bm{w}_{ij}$ is the correlation weight. The obtained $\bm{q}^{'}_{j}\in\mathbb{R}^C$ considers the similarity between the template and search area, and characterizes the global shape information of the target through the max pooling function.

%Therefore, we can obtain the embedded global clues $\bm{q}^{'}_{j}\in\mathbb{R}^d$ for $j$-th point in the search area. We are the champion

\textbf{Template local feature embedding.} To characterize the local similarity between the template and search area, we first obtain the similarity map by computing the cosine distance between them. The similarity function $f_{sim}$ is written as:
\begin{equation}
\setlength{\abovedisplayskip}{1pt}
\setlength{\belowdisplayskip}{1pt}
%\vspace{-5pt}
\bm{s}_{ij} = f_{sim}(\bm{p}_i, \bm{q_j}) = \frac{\bm{p}_i^{\top}\cdot\bm{q_j}}{\|\bm{p}_i\|_2\cdot\|\bm{q}_j\|_2}, \forall \bm{p}_i\in P, \bm{q}_j\in Q
%\vspace{-5pt}
\end{equation}
where $\bm{s}_{ij}$ indicates the similarity between points $i$ and $j$. We then assign each point in the search area with its most similar point in the template, which is written as:
\begin{equation}
\setlength{\abovedisplayskip}{1pt}
\setlength{\belowdisplayskip}{1pt}
\bm{q}_j^{''} = \operatorname{MLP}([\bm{q}_j,\bm{s}_{kj}, \bm{p}_k, \bm{x}_k]), k=\underset{i=1,2,\ldots,N}{\operatorname{argmax}}\{f_{sim}(\bm{p}_i, \bm{q}_j)\}, \forall \bm{q}_j\in Q
\end{equation}
where $k$ indicates the index of the maximum value of similarity, and $\bm{s}_{kj}$, $\bm{x}_k$ are the corresponding maximum value and 3D coordinate, respectively. $[\cdot,\cdot,\cdot,\cdot]$ represents the concatenation operator. We hence obtain the embedded feature $\bm{q}_j^{''}\in\mathbb{R}^{C}$ of $j$-th point in the search area after using MLP. Finally, we concatenate the obtained global and local feature maps to obtain an enhanced feature map $F=\{\bm{f}_j\}_{j=1}^{M}, \bm{f}_j=\operatorname{MLP}[\bm{q}_j^{'}, \bm{q}_j^{''}]$.

%Since we adopt PointNet++~\cite{qi2017pointnet++} as backbone, $\bm{p}_k$ preserves local context of the template point cloud. Note that $\bm{s}_{kj}$ provides the correlation to distinguish whether the point $j$ is similar to the template point $i$. Therefore, after using MLP, we can embed template's local feature into the potential target in the search area.
%after MLP, template's local information is embedded into $\bm{q}_j^{''}$.

\subsubsection{Shape-Aware Feature Learning}
Due to the sparse and incomplete point clouds of the potential target in the search area, we employ shape-aware feature learning to learn dense geometric features of the target, where the dense and complete point clouds of the target can be obtained. It is expected that the learned features from the generated dense point clouds can characterize the geometric structures of the target better.

%In the sparse search area, 3D object shape information contributes to improve the ability to distinguish the potential target from the background. As a result, after the template feature embedding, we consider generating a dense 3D point cloud to characterize the shape information in order to enhance the discrimination of the potential target.

%In a sparse search area, since there are few points on the object, it is difficult to distinguish the target point from the background point. Therefore, we consider using shape information to enhance the discrimination of target. Specifically, we enhance the shape information of the target by enforcing the point features to generate a complete and dense 3D shape. 

\textbf{Dense ground truth processing.} To obtain the dense 3D point cloud ground truth, we first crop and center points lying inside the target's ground truth bounding box in all frames. We then concatenate all cropped and centered points to generate a dense aligned 3D point cloud, denoted by $X=\{\bm{x}_i\}_{i=1}^{2048}$, where $\bm{x}_i$ is the 3D position, and we fix the number of points to 2048 by randomly discarding and duplicating points.

\begin{figure}
	%\vspace{-15pt}
	\centering
	\includegraphics[width=1.0\textwidth]{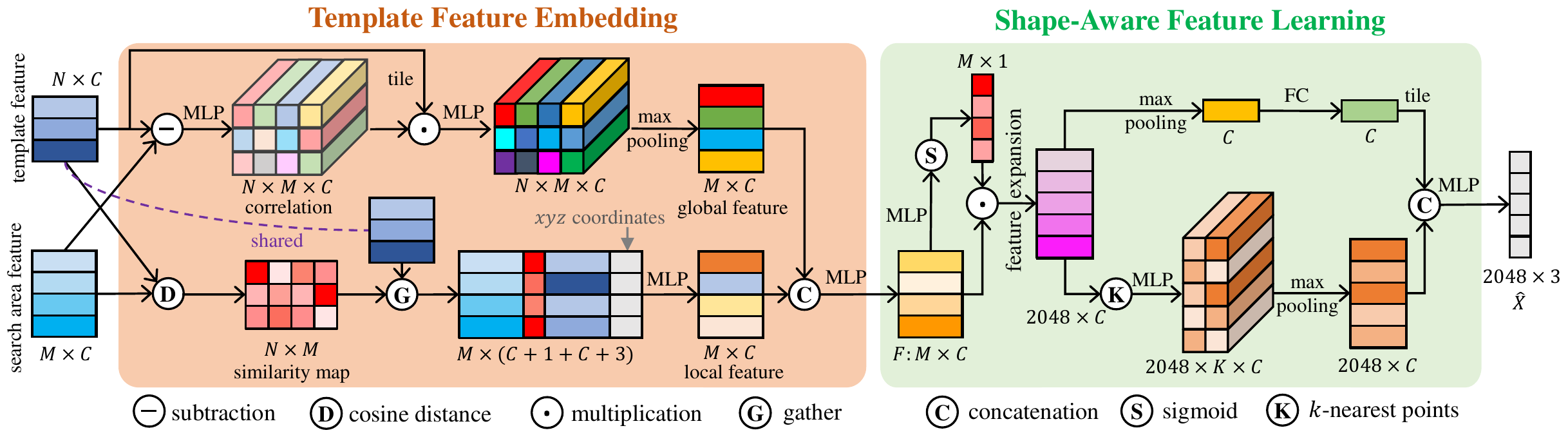}
	%\vspace{-15pt}
	\caption{The architecture of template feature embedding and shape-aware feature learning.}
	\vspace{-15pt}
	\label{fig:siamese_shape}
\end{figure}

%We adopt a two-branch shape generation network, which simultaneously considers the global and local information.
\textbf{Shape information encoding.} We depict the network structure in the right half of Fig.~\ref{fig:siamese_shape}. Suppose the input point feature $\bm{F}\in\mathbb{R}^{M\times C}$ that has been embedded with the template information. Before generating a dense and complete point cloud of the target, we first use a gate mechanism to enhance the feature of the potential target and suppress the background in the search area, which is written as:
\begin{equation}
\setlength{\abovedisplayskip}{1pt}
\setlength{\belowdisplayskip}{1pt}
\bm{F}^{'} = \sigma(\bm{F}\bm{W}^{\top}+b)\circ \bm{F}
\end{equation}
where $\bm{F}^{'}\in\mathbb{R}^{M\times C}$ is the enhanced feature map, $\sigma$ is the \emph{sigmoid} function, and $\circ$ is the element-wise product. Besides, $\bm{W}\in\mathbb{R}^{1\times C}$ is the weight to be learned. It is expected that the potential target can provide more information for generating a dense and complete point cloud of the target. Once we obtain the enhanced feature map, we execute the feature expansion operation~\cite{yu2018pu} to enlarge the feature map from $M\times C$ to $2048\times C$, $i.e.$, the number of points increases from $M$ to $2048$. Then, we capture the global shape information and local geometric structure of the potential target to generate the complete and dense 3D shape. On the one hand, we exploit the max pooling combined with fully connected layers to capture global shape information. On the other hand, we adopt EdgeConv~\cite{wang2018dynamic} to capture local geometric information of the target. After that, we augment the local feature of each point with the global shape information, yielding a new feature map of size $2048\times 2C$. Finally, we adopt MLP to generate 3D coordinates, denoted by $\hat{X}=\{\hat{\bm{x}}_i\}_{i=1}^{2048}$. To train the shape generation network, we follow~\cite{giancola2019leveraging,fan2017point} and use Chamfer distance (CD) loss to enforce the network to generate a realistic 3D point cloud. The Chamfer distance measures the similarity between the generated point cloud and dense ground truth, which is given by:
\begin{equation}
\setlength{\abovedisplayskip}{1pt}
\setlength{\belowdisplayskip}{1pt}
\mathcal{L}_{shape} = \sum_{\bm{x}_i\in X}\underset{\hat{\bm{x}}_j\in \hat{X}}{\min}\|\bm{x}_i-\hat{\bm{x}}_j\|_2^2+\sum_{\hat{\bm{x}}_j\in \hat{X}}\underset{\bm{x}_i\in X}{\min}\|\bm{x}_i-\hat{\bm{x}}_j\|_2^2
\end{equation}
By minimizing the CD loss, we can learn dense geometric features of the potential target in the search area by generating a dense and complete point cloud of the target. Note that shape information encoding is only performed during training and will be discarded during testing. Thus, it does not increase the inference time of object tracking in the test scheme. 

Although SC3D~\cite{giancola2019leveraging} also uses shape completion to encode shape information of the target, the template completion model in SC3D cannot recover complex geometric structures of the potential target well due to limited templates and large variations of the potential target in the search area. However, our method is a complex point cloud generation method that learns the target completion model from the samples of search areas with the gate mechanism to enhance the feature of the potential target and suppress the background in the search area. In addition, the template completion model in SC3D only employs PointNet~\cite{qi2017pointnet} to extract point features of sparse point clouds, while our target completion model constructs a global-local branch to extract global shape features and local geometric features of sparse point clouds.

%It is noted that our shape-aware feature learning is different from SC3D~\cite{giancola2019leveraging}. We perform shape-aware feature learning on the entire search area, while SC3D performs shape completion on the template. Besides, we consider the global and local feature of the point cloud, while SC3D only considers the global feature.

\subsection{Voxel-to-BEV Target Localization Network}\label{sec:center_detection}
In order to avoid using the low-quality proposals on sparse point clouds for target center prediction, we develop a simple yet effective target center localization network without any proposal to improve the localization precision in sparse point clouds.

\subsubsection{Dense BEV Feature Map Generation}
In order to improve the localization precision in sparse point clouds, we utilize the voxelization and max-pooling operation to convert the learned discriminative features of sparse 3D points into the dense bird's eye view (BEV) feature map for the target localization, as shown in the right half of Fig.~\ref{fig:framework}. We first convert the point features of the search area into a volumetric representation by averaging the 3D coordinates and features of the points in the same voxel bin. Then, we apply a stack of 3D convolutions on the voxelized feature map to aggregate the feature of the potential target in the search area, where the voxels lying on the target can be encoded with rich target information. However, in the sparse volume space, due to the large number of empty voxels, the differences between the responses in the voxelized feature map might not be remarkable. Thus, the highest response in the feature map is difficult to distinguish from the low responses, leading to the inaccurate regression of the 3D center of the target, including the $z$-axis center. By performing max-pooling on the voxelized feature map along the $z$-axis, we can obtain the dense BEV feature map, where the low responses in the voxelized feature map can be suppressed. Thus, compared to the voxelized feature map, we can more accurately localize the 2D center of the target with the highest response in the dense BEV feature map. The response of the 2D center ($i.e.$, max-pooling feature along the $z$-axis) in the BEV feature map actually contains the geometric structure information of the potential target while the responses of other points in the BEV feature map do not. In addition, we apply a stack of 2D convolutions on the dense BEV feature map to aggregate the feature so that the potential target can obtain sufficient local information in the BEV feature map. Thus, with the constructed dense BEV feature map, for sparse point clouds, our method can more accurately localize the target center without any proposal.

%Specifically, we first translate all points into the local coordinate system, and then normalize the points in voxel by averaging the coordinates and features of the points.

%To enhance target features, we further apply a stack of 3D convolutions to the volumetric space to aggregate the features so that target's voxel can contain sufficient target information. After that, we compress the 3D volumetric representation into the 2D representation by using max pooling along the $z$-axis, that is, the BEV feature map is obtained. Subsequently, we apply a stack of 2D convolutions to aggregate features so that target's pixel can obtain sufficient target information. Finally, we can obtain a denser BEV feature map that contains rich target information.
%For more details, please refer to the supplementary material.

%\vspace{-5pt}
\subsubsection{Target Localization in BEV}
Inspired by~\cite{Ge2020AFDetAF}, we develop a simple yet powerful network to detect the 2D center and the $z$-axis center based on the obtained dense BEV feature map. As shown in the right half of Fig.~\ref{fig:framework}, it consists of three heads: 2D-center head, offset \& rotation head, and $z$-axis head. The 2D-center head aims to localize 2D center of target on the $x$-$y$ plane, and $z$-axis head regresses the target center of the $z$-axis. Since the 2D center of 2D grid is discrete, we also regress the offset between it and the continuous center. Thus, we use a offset \& rotation head to regress offset plus additional rotation.

\textbf{Target center parameterization.} Given the voxel size $v$ and the range of the search area $[(x_{min}, x_{max}), (y_{min}, y_{max})]$ in $x$-$y$ plane, we can obtain the resolution of the BEV feature map by $H=\lfloor \frac{x_{max}-x_{min}}{v}\rfloor+1$ and $W=\lfloor\frac{y_{max}-y_{min}}{v}\rfloor+1$, where $\lfloor\cdot\rfloor$ is the floor operation. Assuming the 3D center $(x,y,z)$ of the target ground truth, we can compute the 2D target center $c=(c_x,c_y)$ in $x$-$y$ plane by $c_x=\frac{x-x_{min}}{v}$ and $c_y=\frac{y-y_{min}}{v}$. Besides, the discrete 2D center $\tilde{c}=(\tilde{c}_x,\tilde{c}_y)$ is defined by $\tilde{c}_x=\lfloor c_x\rfloor$ and $\tilde{c}_y=\lfloor c_y\rfloor$.

\textbf{2D-center head.} Following~\cite{Ge2020AFDetAF}, we obtain the target center's ground truth $\mathcal{H}\in\mathbb{R}^{H\times W\times 1}$. For the pixel $(i, j)$ in the 2D bounding box, if $i=\tilde{c}_x$ and $j=\tilde{c}_y$, the $\mathcal{H}_{ij}=1$, otherwise $\frac{1}{d+1}$, where $d$ represents the Euclidean distance between the pixel $(i, j)$ and the target center $(\tilde{c}_x, \tilde{c}_y)$. For any pixel outside the 2D bounding box, $\mathcal{H}_{ij}$ is set to 0. In the training phase, we enforce the predicted map $\hat{\mathcal{H}}\in\mathbb{R}^{H\times W\times 1}$ to approach the ground truth $\mathcal{H}$ by using Focal loss~\cite{Lin2020FocalLF}. The modified Focal loss is formulated as:
\begin{equation}
\setlength{\abovedisplayskip}{1pt}
\setlength{\belowdisplayskip}{1pt}
\mathcal{L}_{center}=-\sum_{}^{} \mathbb{I}[\mathcal{H}_{ij}=1]\cdot(1-\hat{\mathcal{H}}_{ij})^{\alpha}\log(\hat{\mathcal{H}}_{ij}) +\mathbb{I}[\mathcal{H}_{ij}\neq 1]\cdot (1-\mathcal{H}_{ij})^{\beta}(\hat{\mathcal{H}}_{ij})^{\alpha}\log(1-\hat{\mathcal{H}}_{ij})
\end{equation}
where $\mathbb{I}(\emph{cond.})$ is the indicator function. If \emph{cond.} is true, then $\mathbb{I}(\emph{cond.})=1$, otherwise 0. Besides, we empirically set $\alpha=2$ and $\beta=4$ in all experiments. %By minimizing the $\mathcal{L}_{heat}$, the heatmap head can accurately regress the 2D center of the object. 

%For the offset head, we predict the offset regression map, which is a square area with radius $r$ around the predicted center of the object.
\textbf{Offset \& rotation head.}
%Since the continuous 2D object center is converted into the discrete one through floor operation, the obtained 2D object center is rough. Therefore, we use the offset head to regress the offset of the continuous ground truth center.
Since the continuous 2D object center is converted into the discrete one by floor operation, we consider regressing the offset of the continuous ground truth center. To improve the accuracy of regression, we consider a square area with radius $r$ around the object center. Here, we also add rotation regression. Given a predicted map $\hat{\mathcal{O}}\in\mathbb{R}^{H\times W\times 3}$, where $3$-dim means the 2D coordinate offset plus rotation, the error of offset and rotation is expressed as:
\begin{equation}
\setlength{\abovedisplayskip}{1pt}
\setlength{\belowdisplayskip}{1pt}
\mathcal{L}_{off}=\sum_{\triangle x=-r}^{r}\sum_{\triangle y =-r}^{r}\left|\hat{\mathcal{O}}_{\tilde{c}+(\triangle x,\triangle y)}-[c-\tilde{c}+(\triangle x, \triangle y ),\theta]\right|
\end{equation}
where $\tilde{c}$ and $c$ mean the discrete and continuous position of the ground truth center, respectively. Besides, $\theta$ indicates the ground truth rotation angle and $[\cdot,\cdot]$ is the concatenation operation. %By minimizing the errors, it is desired that the network can predict the 2D center and rotation more accurately.

\textbf{$z$-axis head.} We directly regress the $z$-axis location of the target center from the BEV feature map. Given a predicted map $\hat{\mathcal{Z}}\in\mathbb{R}^{H\times W\times 1}$, we use $L_1$ loss to compute the error of $z$-axis center by: %$\mathcal{L}_{z}=\left|\hat{\mathcal{Z}}_{\tilde{c}}-z\right|$,
\begin{equation}
\setlength{\abovedisplayskip}{1pt}
\setlength{\belowdisplayskip}{1pt}
\mathcal{L}_{z}=\left|\hat{\mathcal{Z}}_{\tilde{c}}-z\right|
\end{equation}
where $\tilde{c}$ is the discrete object center, and $z$ is $z$-axis center's ground truth.

The final loss of our network is as follows: $\mathcal{L}_{total}=\lambda_{1} \mathcal{L}_{shape}+ \lambda_{2} (\mathcal{L}_{center} + \mathcal{L}_{off}) +\lambda_{3} \mathcal{L}_{z}$,
%\begin{equation}
%\begin{split}
%\mathcal{L}_{total}=\lambda_{1} \mathcal{L}_{shape}+ \lambda_{2} (\mathcal{L}_{heat} + \mathcal{L}_{off}) +\lambda_{3} \mathcal{L}_{z}
%\end{split}
%\end{equation}
where $\lambda_{1}$, $\lambda_{2}$, and $\lambda_{3}$ are the hyperparameter for shape generation, 2D center and offset regression, and $z$-axis position regression, respectively. In the experiment, we set $\lambda_{1}=10^{-6}$, $\lambda_{2}=1.0$, and $\lambda_{3}=2.0$.
%For more details, please refer to the supplementary material.

\section{Experiments}
\subsection{Experimental Settings}
\textbf{Datasets.} For 3D single object tracking, we use KITTI~\cite{Geiger2012AreWR} and nuScenes~\cite{nuscenes2019} datasets for training and evaluation. Since the ground truth of the test set of KITTI dataset cannot be obtained, we follow~\cite{giancola2019leveraging,Qi2020P2BPN} and use the training set to train and evaluate our method. It contains 21 video sequences and 8 types of objects. We use scenes 0-16 for training, scenes 17-18 for validation, and scenes 19-20 for testing. For nuScenes dataset, we use its validation set to evaluate the generalization ability of our method. Note that the nuScenes dataset only labels key frames, so we report the performance evaluated on the key frames.
%Note that the nuScenes dataset only labels key frames, so we apply the official interpolation function to obtain the ground truth of the unlabeled frames between key frames.

%For nuScenes~\cite{nuscenes2019} dataset, we use its training set for training and evaluate our method on its validation set.

%process the nuScenes dataset. We use the training set of nuScenes dataset for training and evaluate the model on its validation set.
%We use it to perform the 3D single object tracking task. 

%It contains 21 video sequences and 8 types of objects. Since the ground truth of the test set cannot be obtained, we follow~\cite{giancola2019leveraging,Qi2020P2BPN} and utilize the training set to train and evaluate our method. Specifically, we use scenes 0-16 for training, scenes 17-18 for validation, and scenes 19-20 for testing.

\textbf{Evaluation metrics.} For 3D single object tracking, we use the \emph{Success} and \emph{Precision} defined in the one pass evaluation (OPE)~\cite{Kristan2016ANP} to evaluate the tracking performance of different methods. \emph{Success} measures the IOU between the predicted and ground truth bounding boxes, while \emph{Precision} measures the error AUC of the distance between the centers of two bounding boxes.

\textbf{Implementation details.} Following~\cite{Qi2020P2BPN}, we set the number of points $N=512$ and $M=1024$ for the template and search area by randomly discarding and duplicating points. For the backbone network, we use a slightly modified PointNet++~\cite{qi2017pointnet++}, which consists of three set-abstraction (SA) layers (with query radius of 0.3, 0.5, and 0.7) and three feature propagation (FP) layers. For each SA layer passed, the points will be randomly downsampled by half. %And the dimension of the output feature is 32. 
For the shape generation network, we generate 2048 points. %Here, we first use feature expansion to generate a dense feature map of neuron size $2048\times 64$. We use the max pooling function combined with two fully connected layers to characterize the global information and use one EdgeConv~\cite{wang2018dynamic} to characterize the local information. 
The global branch is the max pooling combined with two fully connected layers, while the local branch only uses one EdgeConv layer. We use a two layer MLP network to generate 3D coordinates. For 3D center detection, the voxel size is set to 0.3 meters in volumetric space. We stack four 3D convolutions (with stride of 2, 1, 2, 1 along the $z$-axis) and four 2D convolutions (with stride of 2, 1, 1, 2) combined with the skip connections for feature aggregation, respectively. For all experiments, we use Adam~\cite{Kingma2015AdamAM} optimizer with learning rate 0.001 for training, and the learning rate decays by 0.2 every 6 epochs. It takes about 20 epochs to train our model to convergence. %Please refer to the supplementary material for more details.

\textbf{Training and testing.} For training, we combine the points inside the first ground truth bounding box (GTBB) and the points inside the previous GTBB plus the random offset as the template of the current frame. To generate the search area, we enlarge the current GTBB by 2 meters and plus the random offset. For testing, we fuse the points inside the first GTBB and the previous result's point cloud (if exists) as the template. Besides, we first enlarge the previous result by 2 meters in current frame, and then collect the points lying in it to generate the search area.
%%%%%%%%%%%%%%%%%%%%%%%%%%%%%%%%%%%%%%%%%%%%%%%%%%%%%%%%%%%%
  
%Due to n proposals, SC3D achieves the highest performance in the cyclist category with few training samples.
\subsection{Results}
\textbf{Quantitative results.} We compare our method with current state-of-the-art methods, including SC3D~\cite{giancola2019leveraging}, P2B~\cite{Qi2020P2BPN}, and BAT~\cite{zheng2021box}. The quantitative results are listed in Tab.~\ref{tab:results_four}. For the KITTI~\cite{Geiger2012AreWR} dataset, we follow~\cite{giancola2019leveraging,Qi2020P2BPN} and report the performance of four categories, including car, pedestrian, van, and cyclist, and their average results. As one can see from the table, our method is significantly better than other methods on the mean results of four categories. For the car category, our method can even improve the \emph{Success} from 60.5\% (BAT) to 70.5\% (V2B). However, for tracking-by-detection methods that rely on large amounts of training samples, it is difficult to effectively track cyclists with few training samples. Thus, P2B, BAT, and V2B are worse than SC3D in the cyclist category. However, SC3D uses exhaustive search to generate numerous candidate proposals for template matching, so it performs well with few training samples. For the nuScenes~\cite{nuscenes2019} dataset, we directly apply the models, trained on the corresponding categories of the KITTI dataset, to evaluate performance on the nuScenes dataset. Specifically, the corresponding categories between KITTI and nuScenes datasets are Car$\rightarrow$Car, Pedestrian$\rightarrow$Pedestrian, Van$\rightarrow$Truck, and Cyclist$\rightarrow$Bicycle, respectively. It can be seen that our V2B can still achieve better performance on the mean results of all four categories. Due to few training samples on the van category, P2B, BAT, and V2B cannot obtain good generalization ability compared to SC3D in the truck category. The quantitative results on the nuScenes dataset further demonstrate that our V2B has a good generalization ability to adapt to different datasets.

%model on the Car, Pedestrian, Van, and Cyclist categories of the KITTI dataset to evaluate on the 
%As one can see from the table, our method is significantly better than other methods on the mean results of four categories. On the car category, our method can even improve the \emph{Success} from 58.4\% (BAT) to 70.5\% (V2B). In addition, SC3D achieves the highest performance in the cyclist category. Since the training samples of cyclists is far less than the other three categories, it is difficult to effectively track cyclists, especially for those tacking-by-detection methods, such as P2B and our V2B. However, SC3D uses exhaustive search to generate numerous candidate proposals for template matching, so it can perform well with few training samples.

\begin{table}
	\caption{The \emph{Success}/\emph{Precision} of different methods on the KITTI and nuScenes datasets. ``Mean'' indicates the average results of four categories.}
	\label{tab:results_four}
	\centering
	%\resizebox{\textwidth}{!}{
	\begin{tabular}{ccccccc}
		\toprule
		\multirow{2}*{Dataset} & Method & Car & Pedestrian & Van & Cyclist & Mean \\
		&Frame Number & 6424 & 6088 & 1248 & 308 & 14068 \\
		\midrule
		\multirow{4}*{KITTI}&SC3D~\cite{giancola2019leveraging} & 41.3~/~57.9 & 18.2~/~37.8 & 40.4~/~47.0 & {\bf41.5}~/~{\bf70.4} & 31.2~/~48.5 \\
		&P2B~\cite{Qi2020P2BPN} & 56.2~/~72.8 & 28.7~/~49.6 & 40.8~/~48.4 & 32.1~/~44.7 & 42.4~/~60.0\\
		%&Re-Track~\cite{Feng2020ANO} & 58.4~/~73.4 & - & - & - & -\\
		&BAT~\cite{zheng2021box} & 60.5~/~77.7 & 42.1~/~70.1 & {\bf 52.4}~/~{\bf 67.0} & 33.7~/~45.4 & 51.2~/~72.8\\
		&V2B (ours) & {\bf70.5}~/~{\bf81.3} & {\bf48.3}~/~{\bf73.5} & 50.1~/~58.0 & 40.8~/~49.7 & {\bf58.4}~/~{\bf75.2} \\  
		\midrule
		\multirow{2}*{Dataset} & Method & Car & Pedestrian & Truck & Bicycle & Mean \\
		&Frame Number & 15578 & 8019 & 3710 & 501 & 27808 \\
		%\cmidrule{2-7}
		\midrule
		\multirow{4}*{nuScenes}&SC3D~\cite{giancola2019leveraging} & 25.0~/~27.1 & 14.2~/~16.2 & {\bf25.7}~/~{\bf21.9} & 17.0~/~18.2 & 21.8~/~23.1 \\
		&P2B~\cite{Qi2020P2BPN} & 27.0~/~29.2 & 15.9~/~22.0 & 21.5~/~16.2 & 20.0~/~{\bf26.4} & 22.9~/~25.3 \\
		&BAT~\cite{zheng2021box} & 22.5~/~24.1 & {\bf17.3}~/~{\bf24.5} & 19.3~/~15.8 & 17.0~/~18.8 & 20.5~/~23.0\\
		&V2B (ours) & {\bf31.3}~/~{\bf35.1} & {\bf17.3}~/~23.4 & 21.7~/~16.7 & {\bf22.2}~/~19.1 & {\bf25.8}~/~{\bf29.0} \\
		\bottomrule
	\end{tabular}
	%}
	\vspace{-20pt}
\end{table}

\textbf{Quantitative results on sparse scenes.} To verify the effectiveness of our method for object tracking in sparse scenes, we count the performance of SC3D, P2B, BAT, and our V2B in sparse scenes of the KITTI dataset. Specifically, we filter out sparse scenes for evaluation according to the number of points lying in the target bounding boxes in the test set. Specifically, the conditions for the sparse scenes are: $\le150$ (car), $\le100$ (pedestrian), $\le150$ (van), and $\le100$ (cyclist), respectively. For the four categories, the number of selected frames are 3293 (car), 1654 (pedestrian), 734 (van), and 59 (cyclist), respectively. In Tab.~\ref{tab:results_four_sparse}, we report the results of \emph{Success} and \emph{Precision}. As one can see from the table, our V2B achieves the best performance on the mean results of all four categories. The results of the cyclist category are worse due to few training samples. Note that when switching from sparse frames (Tab.~\ref{tab:results_four_sparse}) to all types of frames (Tab.~\ref{tab:results_four}), SC3D and P2B suffer from a performance drop on the mean results of four categories. The worse tracking performance of SC3D and P2B on large amounts of sparse frames leads to the inaccurate template updates on the consecutive dense frames. Thus, SC3D and P2B cannot obtain better tracking performance on the dense frames. Although SC3D uses template shape completion, due to limited template samples and large variations of the potential target in the search area, it cannot accurately recover the complex geometric structures of the target in the sparse frames, which poses challenges on localizing the potential target with sparse points. On the contrary, our V2B employs the proposed shape-aware feature learning module to generate dense and complete point clouds of the potential target for the target shape completion, leading to more accurate localization of the target in the sparse frames. Compared with BAT, our V2B achieves the performance gain of 2\% on the mean results of all four categories from sparse frames to all types of frames. Thus, the comparison results can demonstrate that our V2B can effectively improve the performance of single object tracking in sparse point clouds.

%Thus, our method performs better tracking results in the dense frames for the KITTI dataset.
%As one can see from the table, our V2B outperforms BAT for all four categories, which further demonstrates the effectiveness of our V2B in improving tracking performance in sparse point clouds.
%Note that we run the official code to obtain the results of SC3D, P2B, and BAT.

\begin{table}[h]
	%\vspace{-15pt}
	\caption{Comparison of \emph{Success}/\emph{Precision} of different methods on the {\bf sparse} scenarios.}
	\label{tab:results_four_sparse}
	\centering
	%\resizebox{\textwidth}{!}{
	\begin{tabular}{cccccc}
		\toprule
		Method  & Car &Pedestrian & Van & Cyclist & Mean \\
		Frame Number & 3293 & 1654 & 734 & 59 & 5740\\ 
		\midrule
		SC3D~\cite{giancola2019leveraging}& 37.9~/~53.0 & 20.1~/~42.0 & 36.2~/~48.7 & {\bf 50.2}~/~{\bf 69.2} & 32.7~/~49.4 \\
		P2B~\cite{Qi2020P2BPN} & 56.0~/~70.6 & 33.1~/~58.2 & 41.1~/~46.3 & 24.1~/~28.3 & 47.2~/~63.5 \\
		BAT~\cite{zheng2021box}& 60.7~/~75.5 & 48.3~/~{\bf 77.1} & 41.5~/~47.4 & 25.3~/~30.5 & 54.3~/~71.9  \\
		V2B (ours)& {\bf 64.7}~/~{\bf 77.4} & {\bf 50.8}~/~74.2 & {\bf 46.8}~/~{\bf 55.1} & 30.4~/~37.2 & {\bf 58.0}~/~{\bf 73.2} \\  
		\bottomrule
	\end{tabular}
	%}
	%\vspace{-10pt}
\end{table}

%Compared with Tabs.~\ref{tab:results_four} and~\ref{tab:results_four_sparse}, when switching from sparse frames to all types of frames, SC3D and P2B suffer from a performance drop on the mean results of four categories. Since in the single object tracking method the template in the current frame will be updated with the localization of the target in the previous frame, the inaccurately updated template will lead to poor tracking performance on the consecutive frames. The worse tracking performance of SC3D and P2B on large amounts of sparse frames leads to the inaccurate template updates on the consecutive dense frames. Thus, SC3D and P2B cannot obtain better tracking performance on the dense frames. On the contrary, our V2B-Tracker employs the proposed shape-aware feature learning module to generate dense and complete point clouds of the potential target for the target shape completion, leading to more accurate localization of the target in the sparse frames. Thus, our method performs better tracking results in the dense frames for the KITTI dataset.

%our method can obtain better performance in both sparse and all types of 

\begin{figure}[h]
	\centering
	\includegraphics[width=1.0\textwidth]{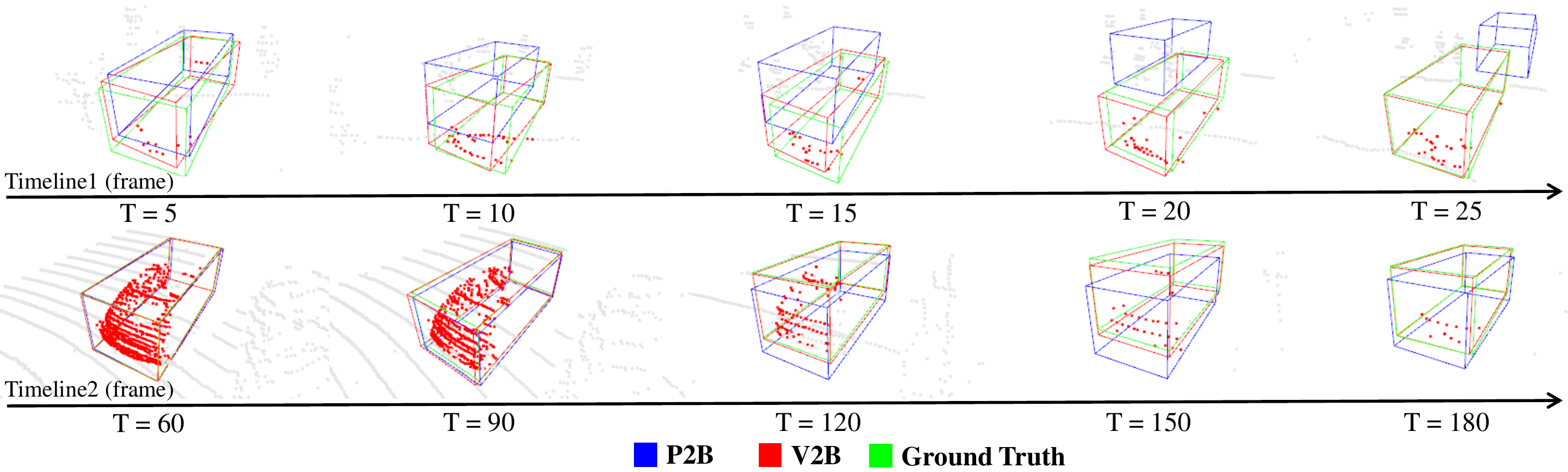}
	%\vspace{-5pt}
	\caption{Visualization results of sparse (the first-row sequence) and dense (the second-row sequence) scenes on the car category. The {\color{green} green} boxes are ground truth bounding boxes. The {\color{red} red} boxes are the objects tracked by our V2B, while the {\color{blue} blue} boxes are the objects tracked by P2B. In addition, we mark the points of cars to {\color{red} red} for better identification.}
	\label{fig:vis_p2b_our}
	\vspace{-15pt}
\end{figure}

\textbf{Visualization results.} As shown in Fig.~\ref{fig:vis_p2b_our}, we plot the visualization results of P2B and our V2B on the car category. Specifically, we plot a couple sparse and dense scenarios on the car category of the KITTI dataset. It can be clearly seen from the figure that compared with P2B, our V2B can track the targets more accurately in both sparse and dense scenes. Especially in sparse scenes, compared with P2B, our V2B can track the targets effectively. The visualization results can demonstrate the effectiveness of our V2B for sparse point clouds.
%For more visualization results, please refer to the supplementary material.

\subsection{Ablation Study}

\begin{wraptable}{r}{6.8cm}
	%\vspace{-10pt}
	\caption{The ablation study results of different components on the car category.}
	\label{tab:ablation}
	\centering
	\begin{tabular}{cc}
		\toprule
		%Module & \emph{Success}/\emph{Precision} \\
		Module & \emph{Success}/\emph{Precision}\\
		\midrule
		w/o template feature & 63.9~/~73.9\\
		local temple & 68.0~/~79.2\\
		global temple & 68.8~/~80.0	\\
		\midrule
		w/o shape information & 67.6~/~78.2\\
		local geometric & 68.6~/~79.3 \\
		global shape & 69.6~/~80.3 \\
		\midrule
		default setting & {\bf 70.5}~/~{\bf 81.3}\\
		\bottomrule
	\end{tabular}
\end{wraptable}

\textbf{Template feature embedding.} We study the impact of template feature embedding on tracking performance. As shown in Tab.~\ref{tab:ablation}, we report the results of the car category in the KITTI dataset. It can be seen that without using the template feature embedding (dubbed ``w/o template feature''), the performance will be greatly reduced from 70.5~/~81.3 to 63.9~/~73.9 by a large margin. In addition, only using the local branch or global branch cannot achieve the best performance. Since template feature embedding builds the relationship between the template and search area, it will contribute to identify the potential target from the background in the search area. Therefore, when the template feature embedding is absent, the performance will be greatly reduced, which further demonstrates the effectiveness of the proposed template feature embedding for improving tracking performance.

\textbf{Shape-aware feature learning.} For sparse point clouds, we further introduce shape-aware feature learning to enhance the ability to distinguish the potential target from the background in the search area. As shown in Tab.~\ref{tab:ablation}, we conduct experiments to demonstrate the effectiveness of the shape information. It can be seen from the table that without using shape-aware feature learning module (dubbed ``w/o shape information''), the performance will reduce from 70.5~/~81.3 to 67.6~/~78.2. In addition, only using the local geometric branch or global shape branch cannot achieve the best performance. Since the shape generation network can capture 3D shape information of the object to learn the discriminative features of potential target so that it can be identified from the search area.

%\begin{wraptable}[h]
\begin{wraptable}{r}{7cm}
	\vspace{-10pt}
	\caption{Comparison of different detection schemes on the {\bf sparse} scenarios.}
	\label{tab:target_localization}
	\centering
	\begin{tabular}{ccc}
		\toprule
		%Module & \emph{Success}/\emph{Precision} \\
		Module & VoteNet & Voxel-to-BEV\\
		\midrule
		Car & 56.9~/~72.0 & {\bf 64.7}~/~{\bf 77.4}\\
		Pedestrian & 35.3~/~62.1 & {\bf 50.8}~/~{\bf 74.2}\\
		Van & 30.7~/~39.0 & {\bf 46.8}~/~{\bf 55.1}\\
		Cyclist & 23.9~/~30.0 & {\bf 30.4}~/~{\bf 37.2}\\
		\bottomrule
	\end{tabular}
	%\vspace{-5pt}
\end{wraptable}
\textbf{Voxel-to-BEV target localization.} Different from SC3D~\cite{giancola2019leveraging} and P2B~\cite{Qi2020P2BPN}, our V2B adopts another route to localize potential target in object tracking. SC3D performs matching between the template and the exhaustive candidate 3D proposals to select the most similar proposal as the target. P2B and BAT use VoteNet~\cite{qi2019deep} to generate 3D target proposals, and select the proposal with the highest score as the target. However, when facing sparse point clouds, it is hard to generate high-quality proposals, so these methods may not be able to track the object effectively. Our V2B is an anchor-free method that does not require generating numerous 3D proposals. Therefore, our method can overcome the above concern. In order to prove this, we use VoteNet instead of voxel-to-BEV target localization to conduct experiments in the KITTI dataset. In Tab.~\ref{tab:target_localization}, we report the results of different detection methods in the sparse scenarios. Likewise, we filter out sparse scenes in the test set for evaluation according to the number of points (refer to the setting of Tab.~\ref{tab:results_four_sparse}). It can be found that the results of VoteNet are lower than that of voxel-to-BEV target localization, which further demonstrates the effectiveness of our method in sparse point clouds.

\textbf{Different voxel sizes.} We compress the voxelized point cloud into a BEV feature map for subsequent target center detection. Since the scope of object tracking is a large area, the size of voxel will affect the size of BEV feature map, thereby affecting the tracking performance. We hence study the impact of different voxel sizes on the tracking performance. Specifically, we consider four sizes, including 0.1, 0.2, 0.3, and 0.4 meters. The \emph{Success}/\emph{Precision} results of the four sizes are 52.5~/~62.2 (0.1m), 68.1~/~79.4 (0.2m), {\bf 70.5}~/~{\bf 81.3} (0.3m), and 69.1~/~80.0 (0.4m), respectively. When the voxel size is set to 0.3 meters, we achieve the best performance. A larger voxel size will increase the sparsity and cause the loss of the target's details. A smaller voxel size will increase the size of the BEV feature map, thereby increasing the difficulty of detecting the center.

\textbf{Template generation scheme.} Following~\cite{giancola2019leveraging,Qi2020P2BPN,zheng2021box}, we study the impact of different template generation schemes on tracking performance. As shown in Tab.~\ref{tab:template}, we report the results of four schemes on the car category in the KITTI dataset. It can be seen from the table that our V2B outperforms SC3D, P2B, and BAT in all schemes by a large margin. Compared with these methods, our V2B can yield stable results on four template generation schemes, which further demonstrates that our method can consistently generate accurate tracking results in all types of frames.
%In the experiment, when the first ground truth and previous result are used to generate the template, we can obtain the highest performance.
%Since SC3D and P2B cannot generate high-quality proposals for sparse scenarios, they cannot perform well in sparse cases.
%In sparse scenarios, it is difficult to generate high-quality proposals for those proposal-based methods, such as SC3D and P2B. However, our method 

\begin{table}[h]
	\vspace{-15pt}
	\caption{The results of different template generation schemes of different methods in the car category.}
	\label{tab:template}
	\centering
	\begin{tabular}{ccccc}
		\toprule
		%Module & \emph{Success}/\emph{Precision} \\
		Scheme & SC3D~\cite{giancola2019leveraging} & P2B~\cite{Qi2020P2BPN} & BAT~\cite{zheng2021box} & V2B (ours) \\
		\midrule
		The First GT & 31.6~/~44.4 & 46.7~/~59.7 & 51.8~/~65.5 & {\bf 67.8}~/~{\bf 79.3} \\
		Previous result & 25.7~/~35.1 & 53.1~/~68.9 & 59.2~/~75.6 & {\bf 70.0}~/~{\bf 81.3}\\
		The First GT \& Previous result & 34.9~/~49.8 & 56.2~/~72.8 & 60.5~/~77.7 & {\bf 70.5}~/~{\bf 81.3}\\
		All previous results & 41.3~/~57.9 & 51.4~/~66.8 & 55.8~/~71.4 & {\bf 69.8}~/~{\bf 81.2}\\
		\bottomrule
	\end{tabular}
	\vspace{-15pt}
\end{table}

\section{Conclusion}
In this paper, we proposed a Siamese voxel-to-BEV (V2B) tracker for 3D single object tracking on sparse point clouds. In order to learn the dense geometric features of the potential target in the search area, we developed a Siamese shape-aware feature learning network that utilizes the target completion model to generate the dense and complete targets. In order to avoid using the low-quality proposals on sparse point clouds for target center prediction, we developed a simple yet effective voxel-to-BEV target localization network that can directly regress the center of the potential target from the dense BEV feature map without any proposal. Rich experiments on the KITTI and nuScenes datasets have demonstrated the effectiveness of our method on sparse point clouds.

%Our V2B consists of the  and the voxel-to-BEV target localization network. For sparse point clouds, the Siamese shape-aware feature learning network exploits the generation network to characterize the shape information of the target in order to improve the discrimination of features. For localizing the target, we convert the sparse 3D space into a dense BEV space and then regress the target's 2D center and the $z$-axis center from it, respectively. Compared with sparse 3D space, regressing target center from the dense 2D feature can improve the tracking performance, especially in the sparse scenes. 
%\appendix

\section*{Acknowledgments}
This work was supported by the National Science Fund of China (Grant Nos. U1713208, 61876084).

%\section{Appendix}

%Optionally include extra information (complete proofs, additional experiments and plots) in the appendix.
%This section will often be part of the supplemental material.

%\bibliographystyle{unsrt}
\bibliographystyle{plain}
\small\bibliography{bibsfile}

\appendix

\section{Overview}
This supplementary material provides more details on network architecture, implementation, and experiments in the main paper to validate and analyze our proposed method. We will release the code after the paper is published.

In Sec.~\ref{sec:network_architecture}, we provide specific network architecture and more details about the target center parameterization for the voxel-to-BEV target localization network. In Sec.~\ref{sec:implementation}, we provide more details about template and search area generation in training and testing. In Sec.~\ref{sec:experiments}, we show more experimental results including quantitative results, visualization, and ablation study.

\begin{figure}[h]
	%\vspace{-15pt}
	\centering
	\includegraphics[width=1.0\textwidth]{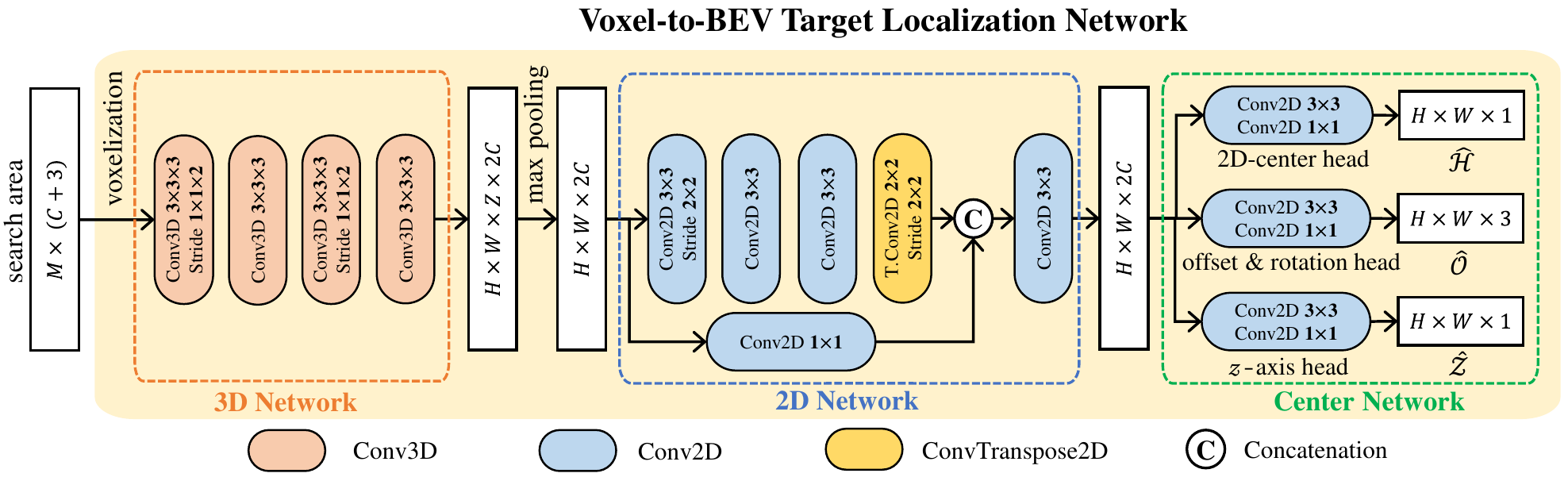}
	%\vspace{-15pt}
	\caption{The architecture of our voxel-to-BEV target localization network.}
	\vspace{-15pt}
	\label{fig:frame_detection}
\end{figure}

\section{Network Architecture}\label{sec:network_architecture}
In this section, we provide specific network architecture used for the voxel-to-BEV target localization network. As shown in Fig.~\ref{fig:frame_detection}, we illustrate the specific network structure. Specifically, we first use the 3D network to aggregate features in the volumetric space. Then, we present the 2D network to aggregate features in the BEV space. After that, we introduce the center network for target localization. Finally, we provide more details on the target center parameterization.

\textbf{3D network.} We use a stack of 3D convolutions to the volumetric space to aggregate the features so that target's voxel can obtain rich target information. As shown on the left side of Fig.~\ref{fig:frame_detection}, we depict the specific structure of the network. Specifically, it consists of four 3D convolutions with the filter sizes 3$\times$3$\times$3. In order to reduce memory consumption, we set the stride of four 3D convolutions to 1$\times$1$\times$2, 1$\times$1$\times$1, 1$\times$1$\times$2, and 1$\times$1$\times$1, respectively. Note that the stride along the $z$-axis is 2, so the feature size in the $x$-$y$ plane will not change. Finally, the 3D network outputs a new feature map with a size of $H\times W\times Z\times 2C$.

\textbf{2D network.} After projecting the voxelized point cloud into the bird's eye view (BEV) space through the max pooling function, we obtain a new BEV feature map with a size of $H\times W\times 2C$. We use a stack of 2D convolutions to construct a shallow encoder-decoder neural network to aggregate the features so that target's pixel can obtain rich target information. We show the specific network structure in the middle part of Fig.~\ref{fig:frame_detection}. Specifically, we first adopt three 2D convolutions (of filter sizes 3$\times$3) and one transposed convolution (of filter size 2$\times$2). And the strides of four 2D convolutions are 2$\times$2, 1$\times$1, 1$\times$1, and 2$\times$2, respectively. Note that the last 2D convolution is the transposed convolution. We also use a concatenated skip connection to fuse low-level and high-level features. Finally, after using a 2D convolution, we obtain a new feature map with a size of $H\times W\times 2C$.

\textbf{Center network.} Since we have known the size (length, width and height) of the object in the template, we only need to regress the target center, offset, and rotation. As shown on the right side of Fig.~\ref{fig:frame_detection}, we illustrate the specific network structure. Specifically, we use three heads to regress the 2D center, offset \& rotation, and $z$-axis location, respectively. For each head, we use two convolutions with filter sizes of 3$\times$3 and 1$\times$1. The output sizes are $H\times W\times 1$ (2D-center head), $H\times W\times 3$ (offset \& rotation head) and $H\times W\times 1$ ($z$-axis head), respectively. Note that for the offset \& rotation head, 3-dim means the 2-dim coordinate offset and 1-dim rotation.

\begin{figure}
	%\vspace{-15pt}
	\centering
	\includegraphics[width=1.0\textwidth]{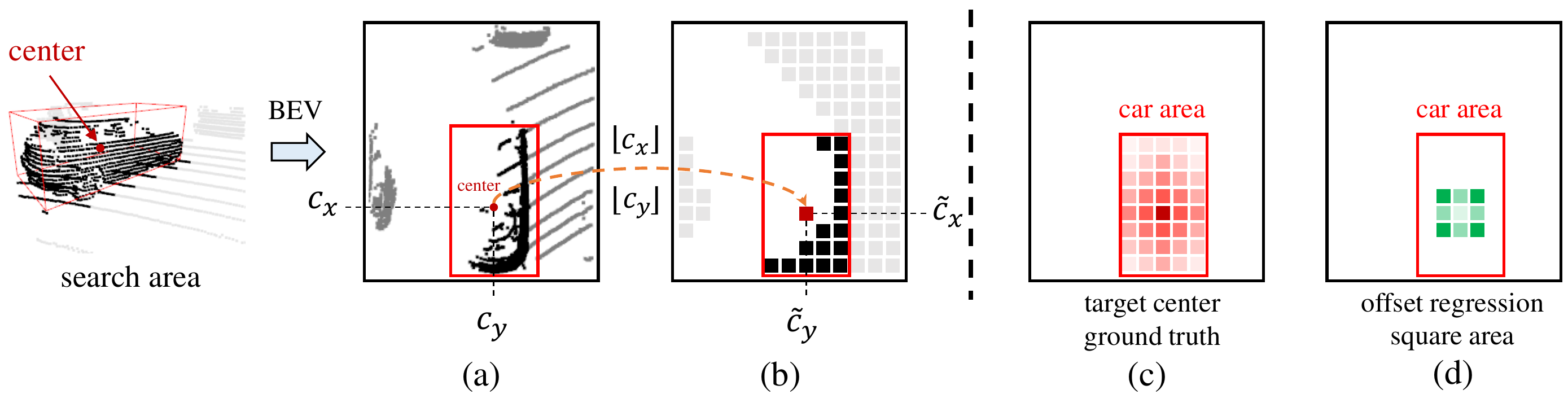}
	%\vspace{-5pt}
	\caption{The details of target center parameterization. (a) represents the continuous coordinates in the BEV space and (b) represents the discrete 2D grid in the BEV space. (c) is the target center ground truth and (d) is the square area used in the offset regression.}
	\vspace{-20pt}
	\label{fig:frame_center}
\end{figure}

\textbf{Target center parameterization.} To obtain the target center in the BEV space, we perform target center parameterization. Assuming the voxel size $v$ and the range of the search area $[(x_{min},x_{max}),(y_{min},y_{max})]$ in the $x$-$y$ plane, we can obtain the resolution of the BEV feature map by:
\begin{equation}
\setlength{\abovedisplayskip}{1pt}
\setlength{\belowdisplayskip}{1pt}
H=\lfloor \frac{x_{max}-x_{min}}{v}\rfloor+1, W=\lfloor\frac{y_{max}-y_{min}}{v}\rfloor+1
\end{equation}
where $\lfloor\cdot\rfloor$ is the floor operation. Given a 3D center $(x,y,z)$ of the target ground truth, we can compute the 2D target center $c=(c_x,c_y)$ in the $x$-$y$ plane by:
\begin{equation}
\setlength{\abovedisplayskip}{1pt}
\setlength{\belowdisplayskip}{1pt}
c_x=\frac{x-x_{min}}{v}, c_y=\frac{y-y_{min}}{v}
\end{equation}
As shown in Fig.~\ref{fig:frame_center}(a), we highlight the target center ($c_x,c_y$) with a {\color{red} red} dot. Note that the current target center is a continuous coordinate. After that, we perform floor operation to obtain the discrete target center $\tilde{c}=$($\tilde{c}_x,\tilde{c}_y$), which is given by
\begin{equation}
\setlength{\abovedisplayskip}{1pt}
\setlength{\belowdisplayskip}{1pt}
\tilde{c}_x=\lfloor c_x\rfloor, \tilde{c}_y=\lfloor c_y\rfloor
\end{equation}
In Fig.~\ref{fig:frame_center}(b), we highlight the discrete target center ($\tilde{c}_x,\tilde{c}_y$) with a {\color{red} red} grid. In this way, we can obtain the 2D grid of the search area.

For the 2D-center head, we follow~\cite{Ge2020AFDetAF} and generate the target center's ground truth $\mathcal{H}\in\mathbb{R}^{H\times W\times1}$ based on the discrete 2D grid. Specifically, for each pixel $(i, j)$ in the 2D bounding box (refer to Fig.~\ref{fig:frame_center}(c)), %if $i=\tilde{c}_x$ and $j=\tilde{c}_y$, the $\mathcal{H}_{ij}=1$, otherwise $\frac{1}{d+1}$
$\mathcal{H}_{ij}$ is defined by:
\begin{equation}
%\setlength{\abovedisplayskip}{1pt}
%\setlength{\belowdisplayskip}{1pt}
%\mathcal{H}_{ij}=\left\{
%\begin{array}{ccl}
%1&      &  \operatorname{if} d = 0\\
%\frac{1}{d+1}     &      & %\operatorname{else}\\
%\end{array} \right.
\mathcal{H}_{ij}=\frac{1}{d+1}
\end{equation}
where $d$ represents the Euclidean distance between the pixel $(i, j)$ and the target center $(\tilde{c}_x, \tilde{c}_y)$. If $i=\tilde{c}_x$ and $j=\tilde{c}_y$, then $\mathcal{H}_{ij}=1$. Besides, for any pixel outside the 2D bounding box, $\mathcal{H}_{ij}$ is set to 0. During training, we enforce the generated map $\hat{\mathcal{H}}\in\mathbb{R}^{H\times W\times 1}$ (refer to the 2D-center head in Fig.~\ref{fig:frame_detection}) to approach the ground truth $\mathcal{H}$ by using Focal loss~\cite{Lin2020FocalLF}, which is given by:
\begin{equation}
\setlength{\abovedisplayskip}{1pt}
\setlength{\belowdisplayskip}{1pt}
\mathcal{L}_{center}=-\sum_{}^{} \mathbb{I}[\mathcal{H}_{ij}=1]\cdot(1-\hat{\mathcal{H}}_{ij})^{\alpha}\log(\hat{\mathcal{H}}_{ij}) +\mathbb{I}[\mathcal{H}_{ij}\neq 1]\cdot (1-\mathcal{H}_{ij})^{\beta}(\hat{\mathcal{H}}_{ij})^{\alpha}\log(1-\hat{\mathcal{H}}_{ij})
\end{equation}
where $\mathbb{I}(\emph{cond.})$ is the indicator function. If \emph{cond.} is true, then $\mathbb{I}(\emph{cond.})=1$, otherwise 0. By minimizing the $\mathcal{L}_{center}$, it is desired that the 2D discrete center can be effectively detected. We empirically set $\alpha=2$ and $\beta=4$ in all experiments.

For the offset \& rotation head, we regress the offset of the continuous ground truth 2D center. Specifically, we consider a square area with radius $r$ (refer to Fig.~\ref{fig:frame_center}(d)) around the object center to improve the accuracy of the offset regression. Note that here we also add rotation regression. Assuming a generated map $\hat{\mathcal{O}}\in\mathbb{R}^{H\times W\times 3}$ (refer to the offset \& rotation head in Fig.~\ref{fig:frame_detection}), the error of the offset and rotation is formulated as:
\begin{equation}
\setlength{\abovedisplayskip}{1pt}
\setlength{\belowdisplayskip}{1pt}
\mathcal{L}_{off}=\sum_{\triangle x=-r}^{r}\sum_{\triangle y =-r}^{r}\left|\hat{\mathcal{O}}_{\tilde{c}+(\triangle x,\triangle y)}-[c-\tilde{c}+(\triangle x, \triangle y ),\theta]\right|
\end{equation}
where $\tilde{c}$ and $c$ mean the discrete and continuous position of the ground truth center, respectively. Besides, $\theta$ indicates the ground truth rotation angle and $[\cdot,\cdot]$ is the concatenation operation. We empirically set $r=2$ in all experiments.

For the $z$-axis head, we directly regress the $z$-axis position of the target center from the BEV feature map. Assuming the generated feature map $\hat{\mathcal{Z}}\in\mathbb{R}^{H\times W\times1}$, the error of the $z$-axis position is written as:
\begin{equation}
\setlength{\abovedisplayskip}{1pt}
\setlength{\belowdisplayskip}{1pt}
\mathcal{L}_{z}=\left|\hat{\mathcal{Z}}_{\tilde{c}}-z\right|
\end{equation}
where $\tilde{c}$ is the discrete target center, and $z$ is $z$-axis center's ground truth. %That is, we only consider the regression value of the target center.

To generate the predicted target center, we first select the position with the highest response on the feature map $\hat{\mathcal{H}}\in\mathbb{R}^{H\times W\times 1}$ as the 2D center on the $x$-$y$ plane. Assuming the selected position is ($i$,$j$), we then obtain the value $\hat{\mathcal{O}}_{ij}\in\mathbb{R}^{3}$ from the feature map $\hat{\mathcal{O}}\in\mathbb{R}^{H\times W\times 3}$ as the value of offset and rotation. After that, we can obtain the continuous target center (${t}_x,{t}_y$) by $t_x=i+\hat{\mathcal{O}}_{ij,0}$, $t_y=j+\hat{\mathcal{O}}_{ij,1}$, where $\hat{\mathcal{O}}_{ij,0}$ and $\hat{\mathcal{O}}_{ij,1}$ are the offsets of the $x$-axis and the $y$-axis. Note that the rotation value $\hat{\mathcal{O}}_{ij,2}$ is performed on the final bounding box. In addition, we obtain the value $\hat{\mathcal{Z}}_{ij}$ from the feature map $\hat{\mathcal{Z}}\in\mathbb{R}^{H\times W\times 1}$ as the target's $z$-axis center.

\section{Implementation Details}\label{sec:implementation}

\begin{figure}
	%\vspace{-15pt}
	\centering
	\includegraphics[width=1.0\textwidth]{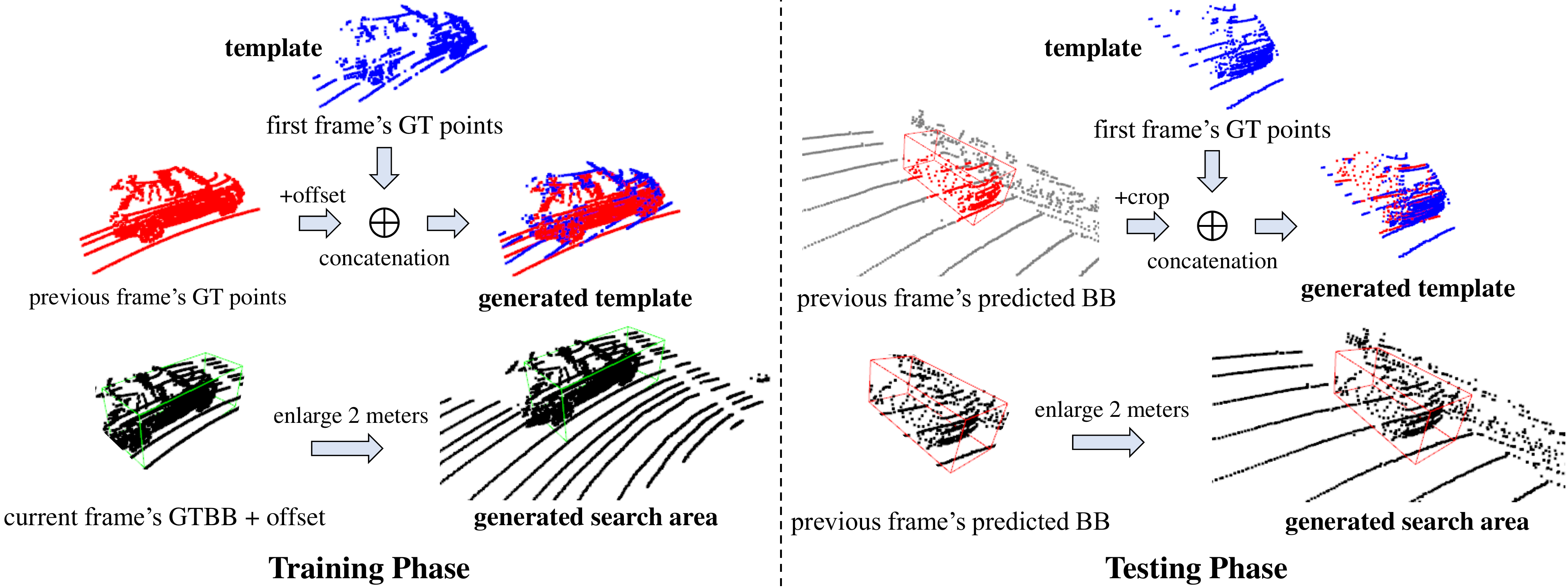}
	\vspace{-5pt}
	\caption{The processing of the template and search area in the training phase and the testing phase, respectively. ``BB'' denotes the bounding box, and ``GT'' denotes the ground truth.}
	\vspace{-15pt}
	\label{fig:template}
\end{figure}

\textbf{Template and search area in training.} Following~\cite{giancola2019leveraging,Qi2020P2BPN}, we adopt the same strategy to generate the template and search area during training. For the current frame, the template is generated by fusing two frames, $i.e.$, the first frame and the previous frame (if exists). As shown in the left half of Fig.~\ref{fig:template}, we combine the points inside the first frame's ground truth bounding box (GTBB) and the points inside the previous frames' GTBB plus the random offset as the template of the current frame. During training, we sample 512 points from the template by discarding and duplicating the points. For the search area, we enlarge the current frame's GTBB by 2 meters and plus the random offset. Likewise, we sample 1024 points from the search area by discarding and duplicating the points.

\textbf{Template and search area in testing.} In the right half of Fig.~\ref{fig:template}, we show the specific process of generating the template and search area. For testing, we fuse the points inside the first frame's ground truth bounding box (GTBB) and the previous result’s point cloud (if exists) as the template of the current frame. For the search area, we first enlarge the previous predicted bounding box by 2 meters in current frame, and then collect the points lying in it to construct the search area. Note that unlike the training phase, we do not apply random offset augmentation to the previous result's point cloud. Besides, we sample 512 points in the template and 1024 points in the search area during testing.

\textbf{Template generation schemes.} In the experiment, we also report the tracking performance of four different template generation schemes. They are dubbed as ``The First GT'', ``Previous result'', ``The First GT \& Previous result'' and ``All previous results'', respectively. ``First GT'' means that we only use the first frame as the template and ``Previous result'' means that we only use the tracked result's point cloud in the previous frame as the template. Thus, ``The First GT \& Previous result'' is a combination of ``First GT'' and `Previous result''. Besides, ``All previous results'' represents that we align all of the previous tracked results' point clouds as a template.

\section{Experiments}\label{sec:experiments}

%\begin{wrapfigure}{r}{0.6\textwidth}
\begin{figure}
	%\vspace{-15pt}
	\centering
	\includegraphics[width=0.7\textwidth]{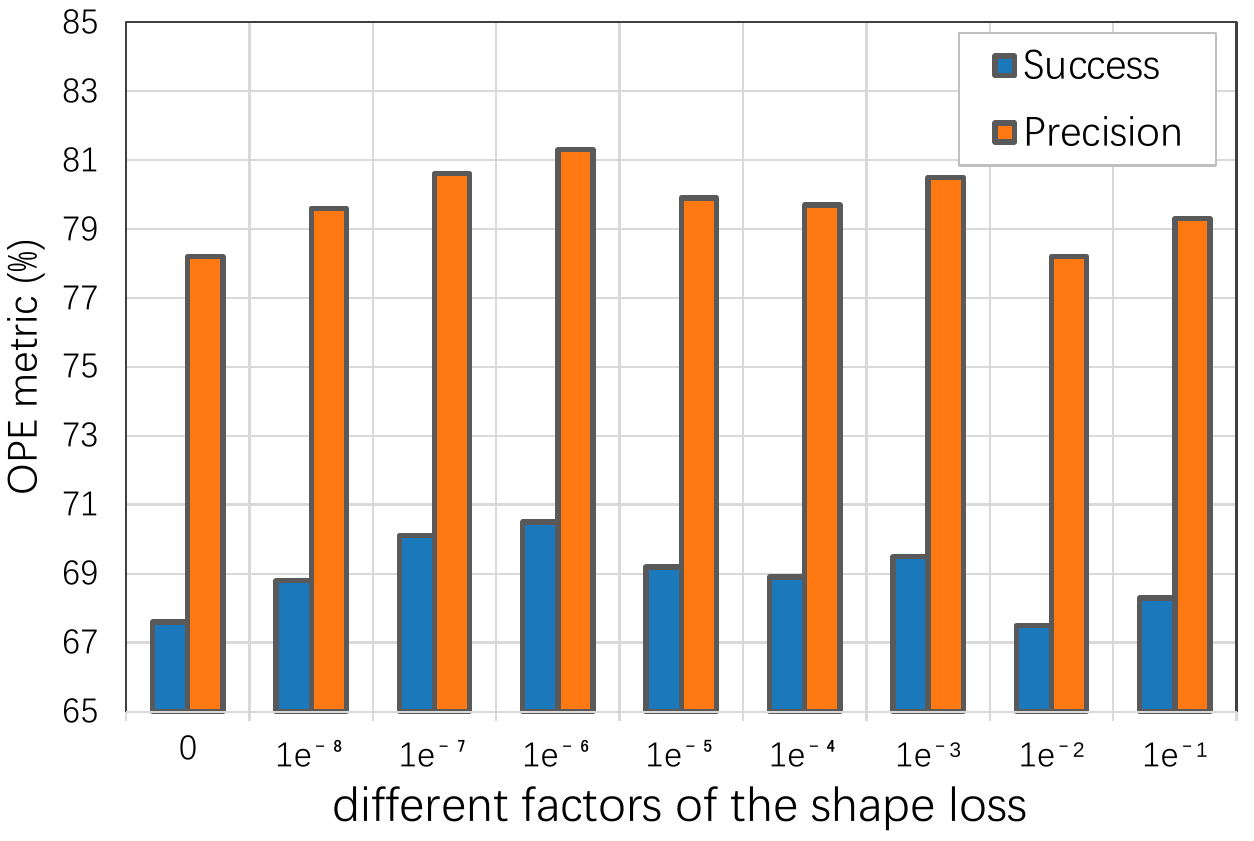}
	\vspace{-5pt}
	\caption{Ablation study results (\emph{Success} and \emph{Precision}) of different factors of the shape generation.}
	\vspace{-15pt}
	\label{fig:shape_loss}
	%\end{wrapfigure}
\end{figure}

\textbf{Different factors of the shape loss.} As shown in Fig.~\ref{fig:shape_loss}, we study the impact of different factors of the shape loss on the tracking performance. It can be seen that when the factor is set to $10^{-6}$, we achieve the best performance. Furthermore, according to the figure, different factors are insensitive to the performance.

\textbf{Quantitative results.} To better validate and analyze the proposed method, we also show the performance of different objects in different point intervals. Specifically, we divide the interval according to the number of points lying in the ground truth bounding boxes in the test set. For large-size categories such as car and van, we set four intervals, including [0, 150), [150, 1000), [1000, 2500), and [2500, +$\infty$). For small-size categories such as pedestrian and cyclist, we set four intervals, including [0, 100), [100, 500), [500, 1000), and [1000, +$\infty$). As shown in Tab.~\ref{tab:results_four_supp}, we report the \emph{Success} and \emph{Precision} of SC3D~\cite{giancola2019leveraging}, P2B~\cite{Qi2020P2BPN}, and our V2B. It can be seen that our method is superior to other methods in terms of the mean of the four categories. In addition, our method is lower than SC3D on the cyclist category. Since there are few training samples on the cyclist category, it will affect the performance of P2B and our method. However, SC3D enumerates exhaustive candidate proposals, so it can achieve higher performance.

\begin{figure}
	%\vspace{-15pt}
	\centering
	\includegraphics[width=1.0\textwidth]{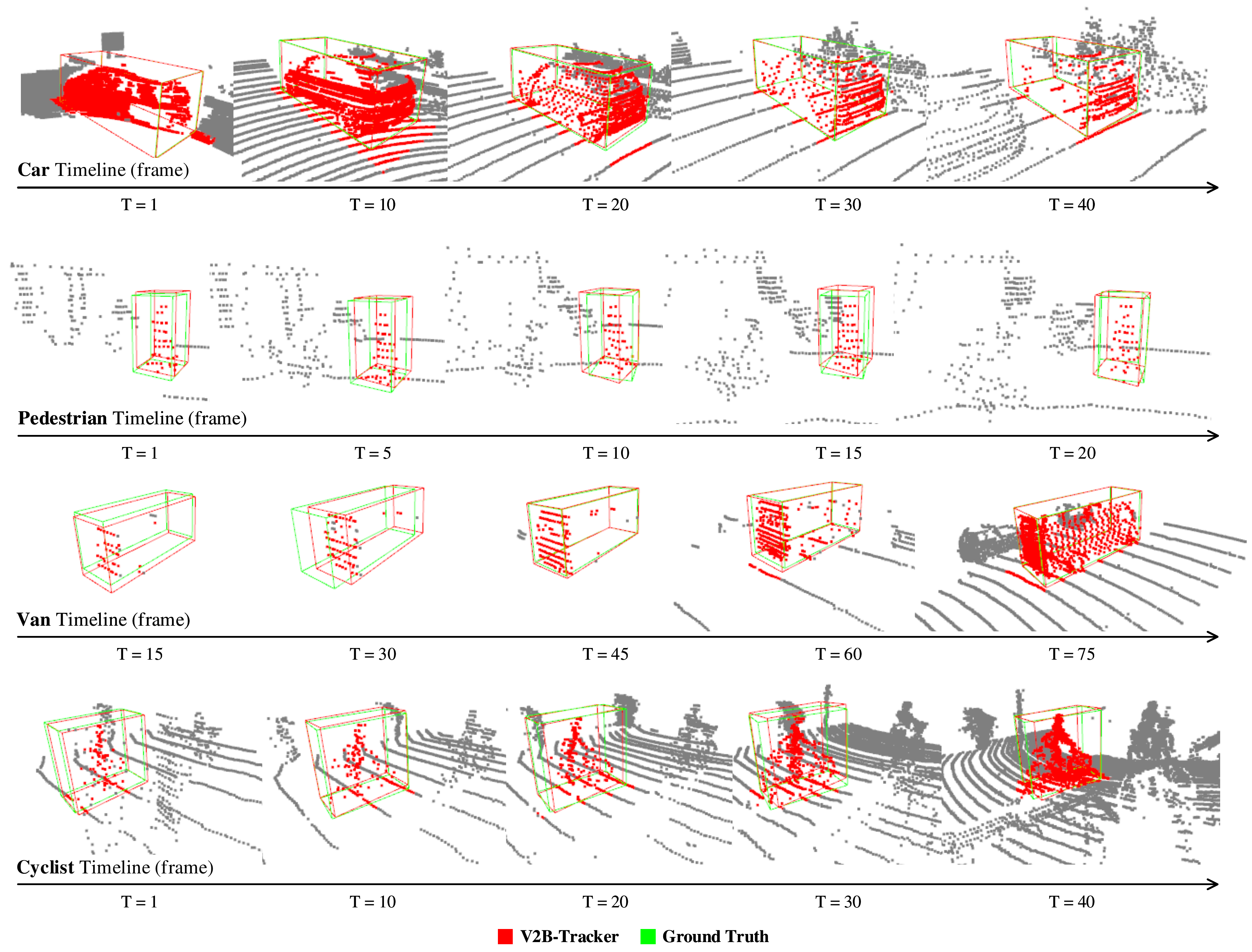}
	\vspace{-10pt}
	\caption{Visualization results of our method. From top to bottom, the visualization results are cars, pedestrians, vans, and cyclists, respectively. The {\color{green} green} boxes are ground truth bounding boxes, and the {\color{red} red} boxes are the predicted bounding boxes of our V2B. Note that we mark the points of the ground truth in {\color{red} red} for better identification.}
	%\vspace{-15pt}
	\label{fig:visualization}
\end{figure}

\begin{figure}
	%\vspace{-15pt}
	\centering
	\includegraphics[width=1.0\textwidth]{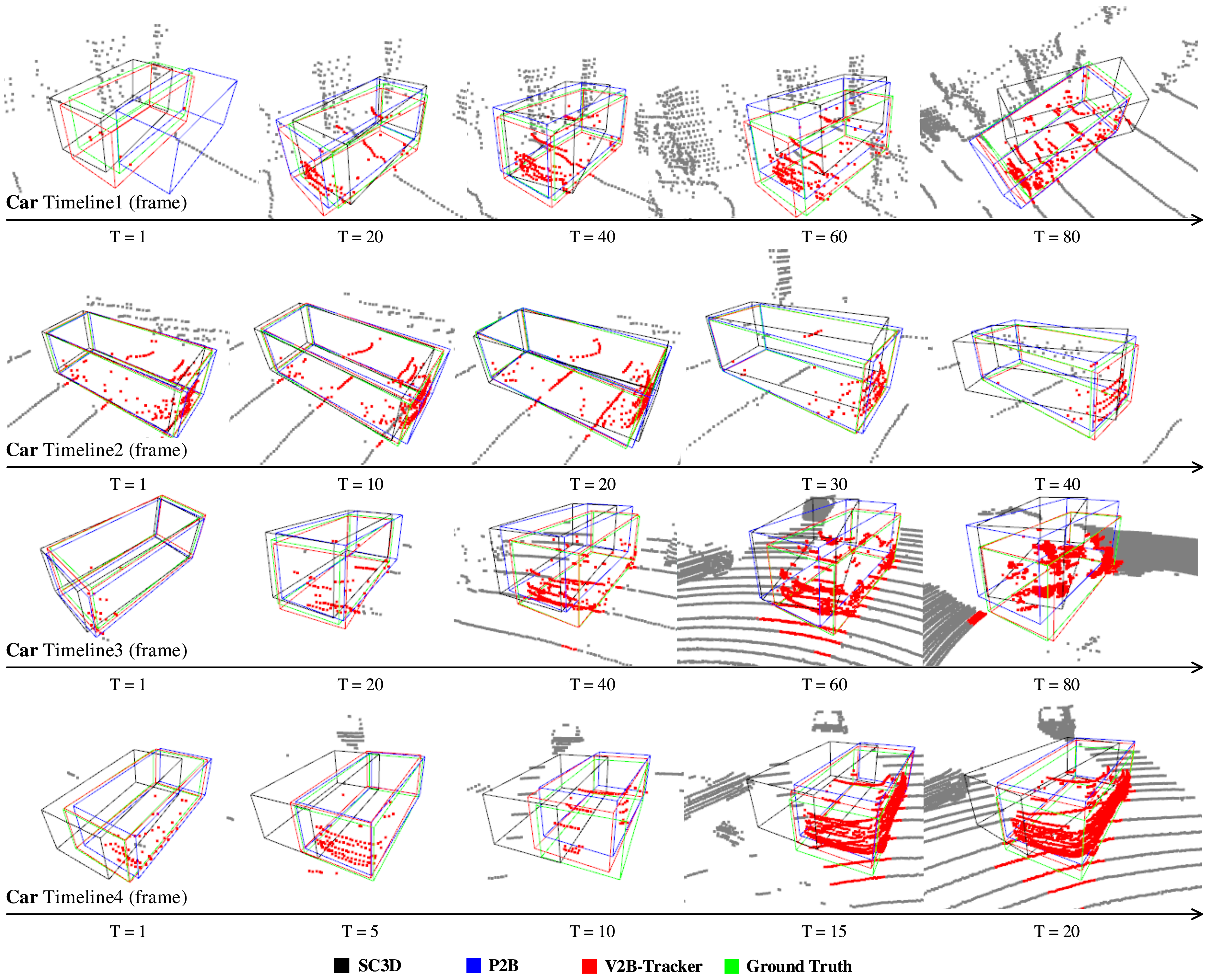}
	\vspace{-10pt}
	\caption{Visualization results of SC3D~\cite{giancola2019leveraging}, P2B~\cite{Qi2020P2BPN}, and our method on the car category. The {\color{green} green} boxes are ground truth bounding boxes. The {\color{black} {\bf black}}, {\color{blue} blue}, {\color{red} red} boxes are the predicted bounding boxes of SC3D, P2B, and our V2B, respectively. Note that we mark the points of the ground truth in {\color{red} red} for better identification.}
	%\vspace{-15pt}
	\label{fig:comparison}
\end{figure}

\textbf{Visualization.} As shown in Fig.~\ref{fig:visualization}, we provide more visualization results of our method for four categories, including car, pedestrian, van, and cyclist. It can be seen that our method can accurately localize targets in both dense and sparse scenes. As shown in Fig.~\ref{fig:comparison}, we also compare the tracking results of SC3D~\cite{giancola2019leveraging}, P2B~\cite{Qi2020P2BPN}, and our method.

\begin{figure}
	%\vspace{-15pt}
	\centering
	\includegraphics[width=1.0\textwidth]{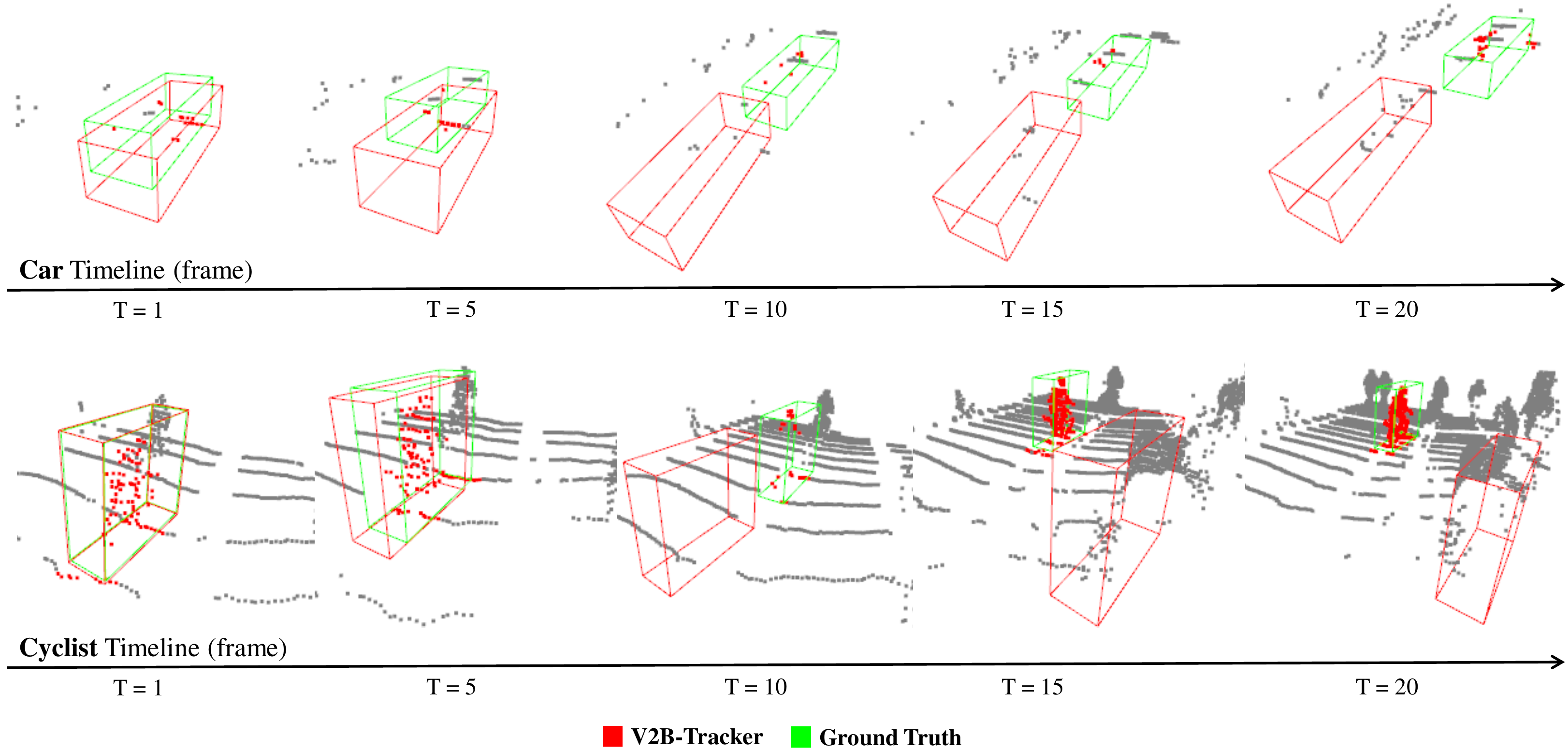}
	\vspace{-10pt}
	\caption{Visualization results of failure cases. The first-row sequence is the result of the car category, and the second-row sequence is the result of the cyclist category. The {\color{green} green} boxes are ground truth bounding boxes, and the {\color{red} red} boxes are the predicted bounding boxes of our V2B. Note that we mark the points of the ground truth in {\color{red} red} for better identification.}
	%\vspace{-15pt}
	\label{fig:failure}
\end{figure}

\textbf{Failure cases.} As shown in Fig.~\ref{fig:failure}, we provide the visualization results of the failure cases. It can be found that our method will fail in extremely sparse scenes. If the tracking fails in the previous frame, a poor-quality search area will be generated for the current frame, which will further affect the tracking results of subsequent frames.

\begin{table}[h]
	\caption{The results of \emph{Success}/\emph{Precision} of different methods at different point intervals. ``Mean'' represents the average results of four categories.}
	\label{tab:results_four_supp}
	\centering
	\resizebox{\textwidth}{!}{
		\begin{tabular}{cccccc}
			\toprule
			Method & Car & Pedestrian & Van & Cyclist & Mean \\
			Total Frame Number & 6424 & 6088 & 1248 & 308 & 14068 \\
			
			\midrule
			Interval & [0, 150) & [0, 100) & [0, 150) & [0, 100) & \\
			Frame Number & 3293 & 1654 & 734 & 59 &  5740\\
			\midrule
			SC3D~\cite{giancola2019leveraging}& 37.9~/~53.0 & 20.1~/~42.0 & 36.2~/~48.7 & {\bf50.2}~/~{\bf69.2} & 32.7~/~49.4 \\
			P2B~\cite{Qi2020P2BPN}& 56.0~/~70.6 & 33.1~/~58.2 & 41.1~/~46.3 & 24.1~/~28.3 & 47.2~/~63.5 \\
			BAT~\cite{zheng2021box}& 60.7~/~75.5 & 48.3~/~{\bf 77.1} & 41.5~/~47.4 & 25.3~/~30.5 & 54.3~/~71.9  \\
			V2B (ours) & {\bf64.7}~/~{\bf77.4} & {\bf50.8}~/~74.2 & {\bf46.8}~/~{\bf55.1} & 30.4~/~37.2 & {\bf58.0}~/~{\bf73.2} \\
			
			\midrule
			Interval & [150, 1000) & [100, 500) & [150, 1000) & [100, 500) & \\
			Frame Number & 2156 & 3112 & 333 & 145 &  5746\\
			\midrule
			SC3D~\cite{giancola2019leveraging}& 36.1~/~53.1 & 17.7~/~38.2 & 38.1~/~53.3 & {\bf44.7}~/~{\bf76.0} & 26.5~/~45.6 \\
			P2B~\cite{Qi2020P2BPN}& 62.3~/~78.6 & 25.1~/~46.0 & 41.7~/~50.5
			& 35.4~/~46.5 & 40.3~/~58.5 \\
			BAT~\cite{zheng2021box} & 71.8~/~83.9 & 45.0~/~71.2 & 44.0~/~51.6 & 41.5~/~52.2 & 54.8~/~74.3 \\
			V2B (ours) & {\bf77.5}~/~{\bf87.1} & {\bf46.8}~/~{\bf72.0} & {\bf51.2}~/~{\bf59.6} & 44.4~/~53.9 & {\bf 58.5}~/~{\bf 76.5} \\
			
			\midrule
			Interval & [1000,2500)  & [500,1000)  & [1000,2500)  & [500,1000) & \\
			Frame Number & 693 & 1071 & 78 & 42 &  1884\\
			\midrule
			SC3D~\cite{giancola2019leveraging}& 33.8~/~48.7
			& 15.0~/~37.1 & 35.9~/~50.3 & 34.9~/~{\bf 69.5} & 23.2~/~42.6 \\
			P2B~\cite{Qi2020P2BPN}& 51.9~/~68.1 & 28.4~/~49.9 & 40.7~/~49.7 & 25.7~/~37.7 & 37.5~/~56.3 \\
			BAT~\cite{zheng2021box} & 69.1~/~81.0 & 35.2~/~61.7 & 50.3~/~61.3 & 34.9~/~48.7 & 48.3~/~68.5 \\
			V2B (ours) & {\bf72.3}~/~{\bf81.5} & {\bf47.2}~/~{\bf74.3} & {\bf61.3}~/~{\bf67.8} & {\bf42.3}~/~52.0 & {\bf 56.9}~/~{\bf 76.2} \\
			
			\midrule
			Interval & [2500,+$\infty$) & [1000,+$\infty$) & [2500,+$\infty$) & [1000,+$\infty$) & \\
			Frame Number & 282 & 251 & 103 & 62 &  698\\
			\midrule
			SC3D~\cite{giancola2019leveraging}& 23.7~/~35.3 & 14.5~/~35.3 & 30.5~/~42.4 & 27.7~/~{\bf 64.2} & 21.8~/~38.9 \\
			P2B~\cite{Qi2020P2BPN}& 43.8~/~61.8 & 27.1~/~49.1 & 33.8~/~39.7 & 24.6~/~34.2 & 34.6~/~51.5 \\
			BAT~\cite{zheng2021box} & 61.6~/~72.9 & 32.6~/~58.6 & 48.2~/~57.9 & 26.7~/~37.9 & 46.1~/~62.4\\
			V2B (ours) & {\bf 82.2}~/~{\bf 90.1} & {\bf 53.8}~/~{\bf 82.6} & {\bf 60.9}~/~{\bf 65.9} & {\bf 41.2}~/~50.4 & {\bf 65.2}~/~{\bf 80.3} \\
			\bottomrule
		\end{tabular}
	}
	\vspace{-15pt}
\end{table}

\end{document}